\documentclass[preprint,3p,12pt,review]{elsarticle}

\usepackage{lineno,hyperref}
\usepackage{longtable}
\usepackage{amsmath,amsfonts,amssymb}
\usepackage[numbers]{natbib}

\usepackage{color}
\definecolor{redcolor}{rgb}{1.0,0.,0.}

\begin{document}

\begin{frontmatter}

\title{Word sense induction using word embeddings and community detection in complex networks}

\author[mymainaddress]{Edilson A. Corr\^ea Jr.}
\author[mymainaddress,cnets]{Diego R. Amancio
\corref{mycorrespondingauthor}
}
\cortext[mycorrespondingauthor]{Corresponding author}
\ead{diego.raphael@gmail.com}

\address[mymainaddress]{Institute of Mathematics and Computer Science, University of S\~ao Paulo (USP)\\
S\~ao Carlos, S\~ao Paulo, Brazil \\}

\address[cnets]{School of Informatics, Computing and Engineering, Indiana University\\
Bloomington, Indiana 47408, USA}

\begin{abstract}
Word Sense Induction (WSI) is the ability to automatically induce word senses from corpora. The WSI task was first proposed to overcome the limitations of manually annotated corpus that are required in word sense disambiguation systems. Even though several works have been proposed to induce word senses, existing systems are still very limited in the sense that they make use of structured, domain-specific knowledge sources. In this paper, we devise a method that leverages recent findings in word embeddings research to generate \emph{context embeddings}, which are embeddings containing information about the semantical context of a word.
In order to induce senses, we modeled the set of ambiguous words as a complex network. In the generated network, two instances (nodes) are connected if the respective \emph{context embeddings} are similar. Upon using well-established community detection methods to cluster the obtained \emph{context embeddings}, we found that the proposed method yields excellent performance for the WSI task. Our method outperformed competing algorithms and baselines, in a completely unsupervised manner and without the need of any additional structured knowledge source.
\end{abstract}

\begin{keyword}
word sense induction \sep language networks \sep complex networks \sep word embeddings \sep community detection \sep word sense disambiguation \sep semantic networks
\end{keyword}

\end{frontmatter}


\section{Introduction}

The Word Sense Induction (WSI) task aims at inducing word senses directly from corpora~\cite{navigli2013semeval}. Since it has been shown that the use of word senses (rather than word forms) can be used to improve the performance of many natural language processing applications, this task has been continuously explored in the literature~\cite{navigli2013semeval,manandhar2010semeval,goyal2014unsupervised}.
In a typical WSI scenario, automatic WSI systems identify the activated sense of a word in a given context, using a variety of features~\cite{navigli2013semeval}. This task is akin to the word sense disambiguation (WSD) problem~\cite{navigli2009word,0295-5075-98-5-58001}, as both induction and disambiguation requires the effective identification of the sense being conveyed. While WSD systems require, in some cases, large corpora of annotated senses, the inductive counterpart (also referred to as unsupervised WSD) does not rely upon any manual annotation~\cite{UREN200614}, avoiding thus the knowledge acquisition bottleneck problem~\cite{gale1992method}.

Analogously to what occurs in supervised disambiguation, WSI techniques based on machine learning represent the state-of-the art, outperforming linguistic-based/inspired methods. Several machine learning methods address the sense identification problem by characterizing the occurrence of an ambiguous word
and then grouping together elements that are similar~\cite{manandhar2010semeval,navigli2013semeval}. The characterization is usually done with the syntactic and semantic properties of the word, and other properties
of the context where it occurs. Once a set of attributes for each occurrence of the ambiguous word is defined, a clustering/grouping method can be easily applied~\cite{manandhar2010semeval,navigli2013semeval}.

Textual contexts are usually represented by vector space models~\cite{Manning:2008:IIR:1394399}. In such models, the context can be represented by the frequency of the words occurring in a given text interval (defined by a window length). Such a representation and its variants are used in several natural language processing (NLP) applications, owing to its simplicity and ability to be used in conjunction with machine learning methods. The integration of machine learning methods and vector space models is facilitated mostly because machine learning methods typically receive structured data as input. Despite of the inherent simplicity of bag-of-word models, in recent years, it has been shown that they yield a naive data representation, a characteristic that might hamper the performance of classification systems~\cite{baroni2014don}. In order to overcome these problems, a novel vector representation -- the \emph{word embeddings} model -- has been used to represent texts~\cite{NIPS2013_5165}. The \emph{word embeddings} representation, also referred to as \emph{neural word embeddings}, are vectors learnt from neural networks in particular language tasks, such as language modeling.
The use of vector representations has led to an improvement in performance of several NLP applications, including machine translation, sentiment analysis and summarization~\cite{taghipour2015semi,baroni2014don,iacobacci2016embeddings,kaageback2015neural}.
%
In the current paper, we leverage the robust representation provided by word embeddings to represent contexts of ambiguous words.


Even though distributional semantic models have already been used to infer senses~\cite{iacobacci2015sensembed}, other potential relevant features for the WSI problem have not been combined with the rich contextual representation provided by the \emph{word embeddings}. For example, it has been shown that the structural organization of the context in bag-of-words models also provides useful information for this problem and related textual problems~\cite{0295-5075-98-1-18002,CORREA2018103}.
For this reason, in this paper, we provide a framework to combine the word embeddings representation with a model that is able to grasp the structural relationship among contexts. More specifically, here we address the WSI problem by explicitly representing texts as a complex network~\cite{perozzi2014inducing}, where words are linked if they are \emph{contextually} similar (according to the word embeddings representation).
By doing so, we found out that the contextual representation is enhanced when the relationship among context words is used to cluster contexts in traditional community detection methods~\cite{FORTUNATO201075,newman2006modularity}. The advantage of using such methods relies on their robustness and efficiency in finding natural groups in highly clustered data~\cite{FORTUNATO201075}. Despite of making use of limited deep linguistic information, our method outperformed several baselines and methods that participated in the SemEval-2013 Task 13~\cite{navigli2013semeval}.


The paper is organized as follows. Section \ref{2} presents some basic concepts and related work. Section \ref{3} presents the details of the proposed WSI method. Section \ref{4} presents the details of the experiments and results. Finally, in Section \ref{5} we discuss some perspectives for further works.

\section{Background and Related Work}\label{2}

The WSI task was originally proposed as an alternative to overcome limitations imposed by systems that rely on sense inventories, which are manually created.
The essential idea behind the WSI task is to group instances of words conveying the same meanings~\cite{navigli2009word}. In some studies, WSI methods are presented as  unsupervised versions of the WSD task, particularly as an effort to overcome the knowledge acquisition bottleneck problem~\cite{gale1992method}. Although some WSI methods have emerged along with the first studies on WSD, a comprehensive evaluation of methods was only possible with the emergence of shared tasks created specifically for the WSI task~\cite{navigli2013semeval,agirre2007semeval,manandhar2010semeval,jurgens2013semeval}.


Several WSI methods use one of the three following methodologies: (i) word clustering; co-occurrence graphs; and (iii) context clustering~\cite{navigli2009word}. Word clustering methods try to take advantage of the semantical similarity between words, a feature that is usually measured in terms of syntactical dependencies~\cite{Sagae,Lin}.
The approach based on co-occurrence graphs constructs networks where nodes represent words and edges are the syntactical relationship between words in the same context (sentence, paragraph or larger pieces of texts). Given the graph representation, word senses are identified via clustering algorithms that use graphs as a source of information~\cite{widdows2002graph,veronis2004hyperlex}. The framework proposed in this manuscript uses the graph representation, however, links are established using a robust similarity measure based on \emph{word embeddings}~\cite{Liu:2015:TWE:2886521.2886657}.
Finally, context clustering methods model each occurrence of an ambiguous word as a context vector, which can be clustered by traditional clustering methods such as Expectation Maximization and $k$-means~\cite{comparacoes}. Differently from graph approaches, the relationship between context words is not explicitly considered in the model.
In~\cite{kaageback2015neural}, the authors explore the idea of context clustering, but instead of using context vectors based on the traditional vector space model (bag-of-words), they propose a method that generates embeddings for both ambiguous and context words.
The method -- referred to as Instance-Context Embeddings (ICE) -- leverages neural word embeddings and  correlation statistics to compute high quality word context embeddings~\cite{kaageback2015neural}.
After the embeddings are computed, they are used as input to the $k$-means algorithm in order to obtain clusters of similar senses. A competitive performance was reported when the method was evaluated in the SemEval-2013 Task 13~\cite{jurgens2013semeval}.
Despite its ability to cluster words conveying the same sense, the performance of the ICE system might be very sensitive to the parameter $k$ in the $k$-means method (equivalently, the number of senses a word can convey), which makes it less reliable in many applications where the parameter is not known a priori.
%


In the present work, we leverage word embeddings to construct complex networks~\cite{0295-5075-98-1-18002,zoran,PhysRevE.88.032910,Breve2013,5871621}. Instead of creating a specific model that generates context embeddings, we use pre-trained embeddings and combine them to generate new embeddings. The use of pre-trained word embeddings is advantageous because these structures store, in a low-cost manner, the semantical contextual information of words trained usually over millions of texts.
Another distinguishing characteristic of our method is that it explores three successful strategies commonly used in WSI.
Firstly, we use semantic information by modeling words via word embeddings. We then make use of complex networks to model the problem. Finally, we use community detection algorithms to cluster instances conveying the same sense. The proposed strategy is also advantageous because
the number of senses do not need to be known a priori, since the network modularity can be used to suggest the number of clusters providing the best partition quality~\cite{newman2006modularity}. The superiority of clustering in networked data over traditional clustering methods has also been reported in the scenario of semantical classification of words~\cite{0295-5075-98-5-58001}.


\section{Overview of the Technique}\label{3}

The proposed method can be divided into three stages: (i) context modeling and context embeddings generation, (ii) network modeling and (iii) sense induction. These steps are described respectively in Sections \ref{sec:cmod}, \ref{sec:compl} and \ref{sec:sensei}.

\subsection{Context modeling and context embeddings generation} \label{sec:cmod}

Several ways of representing the context have been widely stressed by the literature~\cite{navigli2009word}. Some of them consist of using vector space models, also known as bag-of-words, where features are the words  occurring in the context. Other alternative is the use of linguistic features, such as part-of-speech tagging and collocations~\cite{W97-0208}. Some methods even propose to combine two or more of the aforementioned representations~\cite{10.1007/978-981-10-0515-2_8}.

In recent years, a set of features to represent words -- the  word embeddings model -- has become popular. Although the representation of words as vectors has been widely adopted for many years~\cite{navigli2009word}, only recently, with the use of neural networks, this type of representation really thrived. For this reason, from now on word embeddings refer only to the recent word representations, such as \emph{word2Vec} and \emph{GloVe}~\cite{mikolov2013efficient,pennington2014glove}. As in other areas of NLP, word embeddings representations have been used in disambiguation methods, yielding competitive results~\cite{schnabel-EtAl:2015:EMNLP}.

In this work, we decided to model context using word embeddings, mostly because acquiring and creating this representation is a reasonable easy task, since they are obtained in a unsupervised way. In addition, the word embeddings model has been widely reported as the state-of-the art word representation~\cite{zhang-EtAl:2014:P14-11}.
First introduced in~\cite{bengio2003neural}, the neural word embeddings is a distributional model in which words are represented as continuous vectors in an ideally semantic space.
In order to learn these representations, \cite{bengio2003neural} proposed a feed-forward neural network for language modeling that simultaneously learns a distributed representation for words and the probability function for word sequences (i.e., the ability to predict the next word given a preceding sequence of words).
Subsequently, in~\cite{collobert2008unified}, the authors adapted this concept into a deep neural architecture, which has been applied to several NLP tasks, such as part-of-speech tagging, chunking, named entity recognition, and semantic role labeling~\cite{collobert2008unified,collobert2011natural}.

A drawback associated to the architectures devised in~\cite{bengio2003neural,collobert2008unified} is their high computational cost, which makes them prohibitive in certain scenarios. To overcome such a complexity, in~ \cite{mikolov2013efficient,mikolov2013distributed}, the authors proposed the \emph{word2vec} representation. The \emph{word2vec} architecture is similar to the one created in~\cite{bengio2003neural}. However, efficient algorithms were proposed so as to allow a fast training of word embeddings. Rather than being trained in the task of language modeling, two novel tasks were created to evaluate the model: the prediction of a word given its surrounding words (continuous bag-of-words) and the prediction of the context given a word (skipgram).

The word embeddings produced by \emph{word2vec} have the ability to store syntactic and semantic properties~\cite{mikolov2013distributed}. In addition, they have geometric properties that can be explored in different ways. An example is the \emph{compositionality} property, stating that larger blocks of information (such as sentences and paragraphs) can be represented by the simple combination of the embeddings of their words~\cite{mikolov2013efficient,mikolov2013distributed}.
In this work, we leverage this property to create what we define as \emph{context embeddings}. More specifically, we represent an ambiguous word by combining the embeddings of all words in its context (neighboring words in a window of size $w$) using simple operations such as addition.

\begin{figure}[ht]
	   \centering
       \includegraphics[width=0.7\textwidth]{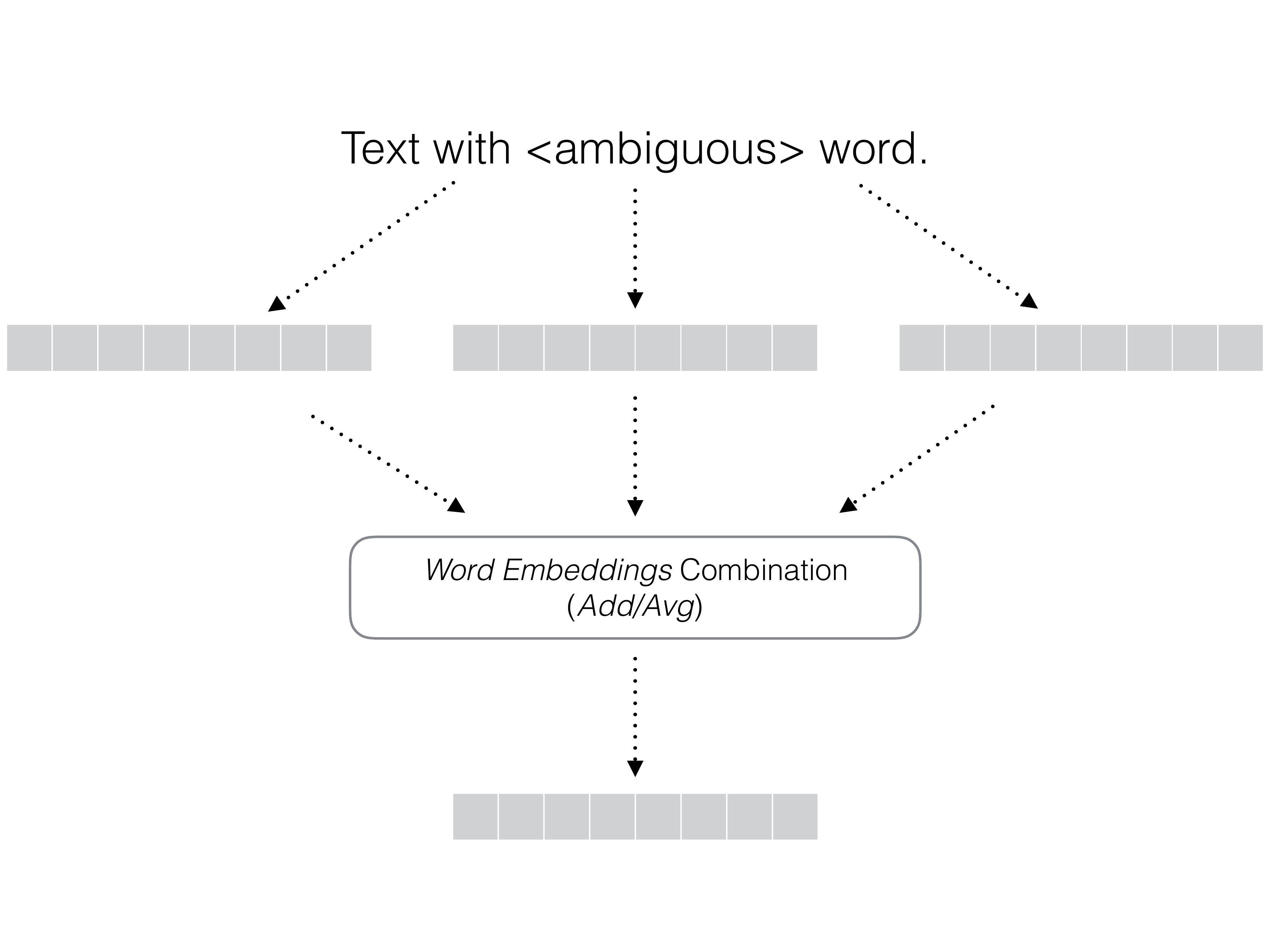}
       \caption{Example illustrating how the context can be characterized from individual word embeddings. Given the word vectors representing the word appearing in the context, we combine those vectors to obtain a single embedding representing the context around the ambiguous word.}
       \label{fig:fig1}
\end{figure}

Figure~\ref{fig:fig1} shows a representation of the process of generating the embeddings of a given occurrence of an ambiguous word. In the first step, we obtain each of the word vectors representing the surrounding words. Particularly, in the current study, the embeddings were obtained from the study conducted in~\cite{mikolov2013efficient,mikolov2013distributed}. The method used to obtain the embeddings is the \emph{word2vec} method, in the skipgram variation~\cite{mikolov2013efficient}. The training phase was performed using \emph{Google News}, a corpus comprising about $100$ billion words. As proposed in~\cite{mikolov2013efficient,mikolov2013distributed}, the parameters for obtaining the methods were optimized considering semantical similarity tasks.
After obtained individual embeddings representing each word in the considered context, such structures are combined into a single vector, which is intended to represent and capture the semantic features of the context around the target word. Here we adopted two distinct types of combination: by (i) addition; and (ii) averaging.

%
%
%

Given the occurrence of the word $w_i$ in a context ($\mathbf{c}_i$) comprising $\omega$ words surrounding $w_i$, i.e. $\mathbf{c}_i$=$[w_{i-\omega/2}$,$\ldots$,$w_{i-1}$,$w_i$, $w_{i+1}$\ldots$w_{i+\omega/2}]^\intercal$, the context embedding ($\mathbf{c}_i$) of $w_i$ obtained from addition is
\begin{equation}
	\mathbf{c}_i = \sum\limits_{\substack{j=-\omega/2 \\ j\neq0}}^{+\omega/2} \mathbf{w}_{i+j},
\end{equation}
where $\mathbf{w}_j$ is the embedding of the $j$-th word in $\mathbf{c}_i$. In other words, the context of a word is given by the composition of the semantic features (word embeddings) associated to the neighboring words. This approach is hereafter referred to as CNN-ADD method.

In the average strategy, a normalizing term is used. Each dimesion of the embedding is divided by the number of words in the context set. Let $l = |\mathbf{c}_i|$ be size of the context.  The average context embedding is defined as:
\begin{equation}
    \mathbf{c}_i = \sum\limits_{\substack{j=-\omega/2 \\ j\neq0}}^{+\omega/2} \frac{\mathbf{w}_{i+j}}{l}.
\end{equation}
This approach is hereafter referred to as CNN-AVG method.

\subsection{Modeling context embeddings as complex networks} \label{sec:compl}

Modeling real-valued vectors into complex networks is a task that can be accomplished in many ways.
Here we represent the similarity between contexts as complex networks, in a similar fashion as it has been done in previous works modeling language networks~\cite{perozzi2014inducing}. While in most works two words are connect if they are similar according to specific criteria, in the proposed model two context vectors are linked if the respective context embeddings are similar. Usually, two strategies have been used to connect nodes.
In the $k$-NN approach, each node is connected to the $k$ nearest (i.e. most similar) nodes. Differently, in the $d$-proximity method, a distance $d$ is fixed and each node is connected to all other nodes with a distance equal or less than $d$~\cite{perozzi2014inducing}.

%

%
%
%
%

\begin{figure}[ht]
	   \centering
       \includegraphics[width=0.65\textwidth]{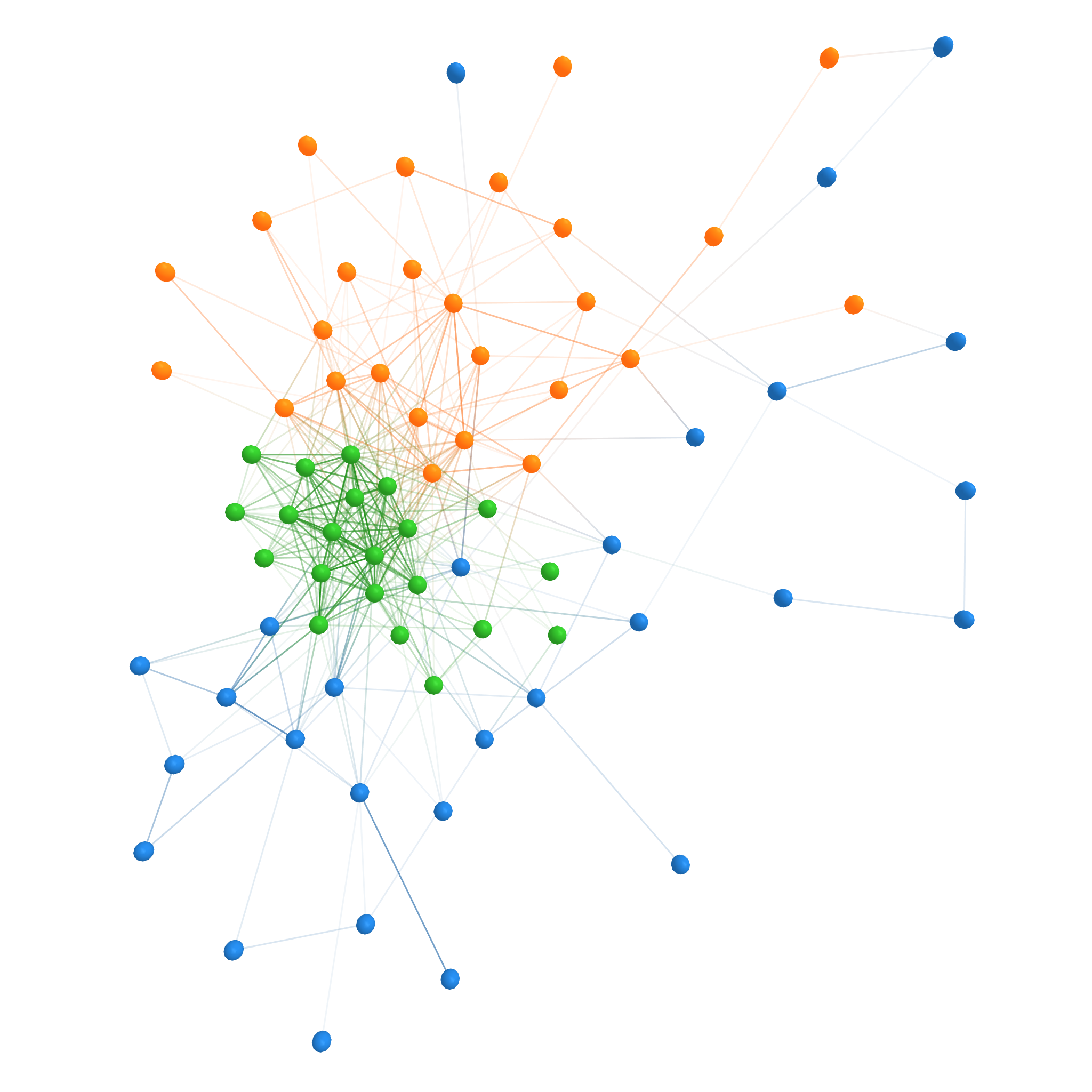}
       \caption{Example of network obtained from the proposed model using $\omega=10$ for the CN-ADD model. Each distinct color represents a different sense induced for the word ``add''. The visualization was obtained with the \emph{networks3d} software~\cite{silva2016using}.}
       \label{fig1}
\end{figure}

In this work, similar to the approach adopted in~\cite{perozzi2014inducing}, we generate complex networks from context embeddings using a $k$-NN approach.
We have chosen this strategy because the network becomes connected with low values of $k$, thus decreasing the complexity of the generated networks. In addition, it has been shown that the $k$-NN strategy is able to optimize the modularity of the generated networks~\cite{perozzi2014inducing}, an important aspect to our method. Both euclidean and cosine were used as distance measurements. In the euclidean case, the inverse of the distances was used as edges weight. In Figure \ref{fig1}, we show the topology of a small network obtained from the proposed methodology for the word ``add''. To construct this visualization, we used $\omega=10$ in the CNN-ADD model. Note that an evident separation among the three distinct senses.

\subsection{Sense induction} \label{sec:sensei}

Once the context embedding network is obtained, the Louvain community detection method~\cite{blondel2008fast} is applied to identify communities. Given the communities produced by the method, we define each community as a induced word sense. We have chosen the Louvain method because it is known to maintain reasonable computational costs~\cite{silva2016using} while maximizing the modularity~\cite{newman2006modularity}. We also have decided  to use  this method because it does not need any additional parameter definition to optimize the modularity function.

\section{Corpora Description}\label{4}

In this section, we present the Semeval-2013 corpora used to evaluate our method. The pre-trained word embeddings used here is also presented.






\subsection{SemEval-2013 Task 13 corpus} The SemEval-2013 data comprises $50$ words. The number of instances of each word ranges between $22$ and $100$ instances. The dataset encompasses $4,664$ instances that were drawn from the \emph{Open American National Corpus}. Each instance is a short piece of text surrounding an ambiguous word that came from a variety of literary genres. The instances were manually inspected to ensure that ambiguous words have at least one interpretation matching one of the WordNet senses.

Following the SemEval-2013 Task 13 proposal~\cite{jurgens2013semeval}, we applied a two-part evaluation setting. In the first evaluation, the induced senses are converted to WordNet 3.1 senses via a mapping procedure and then these senses are used to perform WSD. The output of WSD is evaluated according to the following three aspects:
\begin{enumerate}

\item \emph{Applicability}: this aspect is used to compare the set of senses provided by the system and the gold standard. The applicability criteria, in this context, is measured with the traditional Jaccard Index, which reaches its maximum value when the set of obtained senses and the gold standard are identical.

\item \emph{Senses ranking}: the set of applicable senses for an ambiguous word might consider a different degree of applicability for distinct senses. For this reason, in addition to only considering which senses are applicable, it is also important to probe if the rank of importance assigned for the senses follows the rank defined by the gold standard. The agreement in applicability importance is measured using the positionally-weighted Kendall’s $\tau$ ($K_\delta^{sim}$)~\cite{jurgens2013semeval}.

\item \emph{Human agreement}:  this measurement considers the WSI task as if it were tackled in the information retrieval scenario. In other words, the context of an ambiguous word is a query looking for all senses of the word. The expected retrieved information is the set of all applicable senses, which should be scored and ranked according to the applicability values of the word senses. This criterium was measured using the traditional Normalized Discounted Cumulative Gain (WNDCG) metric, as suggested by the literature~\cite{jurgens2013semeval}.


\end{enumerate}
All above measurements generate values between 0 and 1, where 1 means total agreement with the gold standard. As suggested in similar works, the final score is defined using the F1 measure between each of the objective's measure and the recall~\cite{jurgens2013semeval}. In this case, the recall measures the average score for each measure across all instances, even the ones that were not labeled by the WSD system.

In the second evaluation, the induced senses are compared with a sense inventory through clustering comparisons. In this case, the WSI task is considered as a clustering task and, because each word may be labeled with multiple senses, fuzzy measures are considered.
%
In~\cite{jurgens2013semeval}, the authors propose the use of the  following fuzzy measures:
\begin{enumerate}

\item \emph{Fuzzy B-Cubed}: this measurement summarizes the performance per instance providing an estimate of how well the WSI system would perform on a new corpus with a similar sense distribution.

\item \emph{Fuzzy Normalized Mutual Information}: this index measures the quality of the produced clusters based on the gold standard. Differently from the Fuzzy B-Cubed score, the Fuzzy Normalized Mutual Information is measured at the cluster level, giving an estimate of how well the WSI system would perform independently of the sense distribution of the corpus.

\end{enumerate}

\subsection{Word embeddings} The pre-trained word embeddings\footnote{code.google.com/archive/p/word2vec/} used in this study was trained as a part of the \emph{Google News} dataset, which is composed of approximately $100$ billion words. The model consists of three million distinct words and phrases, where each embedding is made up of $300$ dimensions. All embeddings were trained using the \emph{word2vec} method~\cite{mikolov2013efficient,mikolov2013distributed}.

\section{Results and discussion}

Here we analyze the performance of the proposed methods (section \ref{sec:perf}). In Section \ref{sec:par}, we study the influence of the parameters on the performance of the methods based on complex network created from word embeddings.


\subsection{Performance analysis} \label{sec:perf}

The results obtained by our model were compared with four baselines: (1) One sense, where all instances are labeled with the same sense; (2) 1c1inst, where each instance is defined as a unique sense; (3) SemCor MFS, where each instance is labeled with the most frequent sense of the lemma in the SemCor corpus; and (4) SemCor Ranked Senses, where each instance is labeled with all possible senses for the instance lemma, and each sense is ranked based on its frequency in the SemCor corpus. We also compared our method with the algorithms that participated in the SemEval-2013 shared task.
More specifically, in this task, nine systems were submitted by four
different teams. The AI-KU team submitted three WSI systems based on lexical substitution~\cite{baskaya2013ai}. The University of Melbourne (Unimelb) team submitted two systems based on a Hierarchical Dirichlet Process~\cite{lau2013unimelb}. The University of Sussex (UoS) team submitted two systems relying on  dependency-parsed features~\cite{hope2013uos}. Finally, the La Sapienza team submitted two systems based on the Personalized Page Rank applied to the WordNet in order to measure the similarity between contexts~\cite{agirre2009personalizing}. 

In the proposed method, considering the approaches to generate context embeddings, the general parameter to be defined is the context window size $\omega$. We used the values $\omega = \{1, 2, 3, 4, 5, 7, 10\}$ and the full sentence length. In the network modeling phase, context embeddings are transformed into networks. No parameters are required for defining the \textit{fully-connected} model that generates a fully connected embeddings network. In the $k$-NN model, however, the $k$ value must be specified. We used $k = \{1, 5, 15\}$.



Testing all possible combinations of parameters in our method resulted in $95$ different systems. For simplicity's sake,   only the systems with best performance in the evaluation metrics are discussed in this section. \textcolor{black}{Additional performance results are provided in the Supplementary Information.}
In the following tables the proposed models will be presented by acronyms that refer to the context features used: CN-ADD (Addition) or CN-AVG (Average). CN-ADD/AVG denotes that both systems displayed the same performance. When the $\omega$ column is empty, the full context ({i.e. the full sentence}) was used. Otherwise, the value refers to the context window. The $k$ column refers to the value of the parameter $k$ in the $k$-NN approach used to  create the networks. When $k$ is empty, {the \textit{fully-connected} model was used; otherwise, the value refers to the connectivity of the $k$-NN network.}

Three major evaluations were carried out. In the first evaluation, methods were compared using all instances available in the shared task. The obtained results for this case are shown in Table~\ref{tab1}. Considering the detection of which senses are applicable (see Jacc. Ind. column), our best methods outperformed all participants of the shared task, being only outperformed by the SemCor MFS method, a baseline known for its competitiveness~\cite{Agirre:2009:KWS:1661445.1661686}. Considering the criterium based on  senses rank (as measured by the positionally-weighted Kendall’s $\tau$ ($K_\delta^{sim}$)), our best methods also outperformed all competing systems, including the baselines. In the quantification of senses applicability (WNDCG index), our best methods are close to the participants; however, it is far from the best baseline (SemCor Ranked). Considering the cluster evaluation metrics, our method did not overcome the best baselines, but the same occured to all  participants of the SemEval task. Still, the proposed method outperformed various other methods in the clusters quality, when considering both Fuzzy NMI and Fuzzy B-Cubed criteria.
It is interesting to note that, in this case, the best results were obtained when the fully (weighted) connected network was used to create the networks. In other words, the consideration of all links, though more computationally expensive, seems to allow a better discrimination of senses in this scenario.
\begin{table*}[h]
\centering
\caption{Performance of our best methods evaluated using all instances available in the shared task. The best results are highlighted in bold. Note that, for several criteria, the CN-based method outperformed other traditional approaches.}
\begin{tabular}{@{\extracolsep{4pt}}cccccccc}
\hline
 &  &  & \multicolumn{3}{c}{ WSD F1} & \multicolumn{2}{c}{Cluster Comparison} \\
\cline{4-6} \cline{7-8}
System & $\omega$ & $k$ & Jaccard & $K_\delta^{sim}$ & WNDCG & Fuzzy NMI & Fuzzy B-Cubed \\
\hline
CN-ADD/AVG 	& 10 	& - & \textbf{0.273} & \textbf{0.659} & 0.314 & 0.052 & 0.452 \\
CN-ADD/AVG 	& 5 		& - & 0.266 & 0.650 & \textbf{0.316} & 0.056 & 0.457 \\
CN-ADD 		& 2 		& - 	& 0.252 & 0.588 & 0.293 & \textbf{0.061} & 0.373 \\
CN-ADD/AVG 	& 4 		& 1 & 0.235 & 0.634 & 0.294 & 0.039 & \textbf{0.485} \\

\hline
One sense 	& - 	& - 	& 0.192 	& 0.609 	& 0.288 	& 0.0 	& 0.623 \\
1c1inst 	& - 	& - 	& 0.0 	& 0.0 	& 0.0 	& 0.071 	& 0.0 \\
SemCor MFS 	& - 	& - 	& 0.455 	& 0.465 	& 0.339 	& - 		& - \\
SemCor Ranked& - 	& - 	& 0.149 	& 0.559 	& 0.489 	& - 		& - \\
\hline
\end{tabular}
\label{tab1}
\end{table*}

In the second evaluation, only instances labeled with just one sense were considered. The obtained results are shown in Table \ref{tab2}. Considering F1 to evaluate the sense induction performance, our method outperformed all baselines, but it could not outperform the best participants methods. In the cluster evaluation, conversely, our best method displayed the best performance when compared to almost all other participants. Only two methods (One Sense and SemCor MFS) outperformed our CN approach when considering the instance performance evaluation (as measured by the Fuzzy B-Cubed index). Regarding the best $k$ used to generate networks, we have found that, as in the previous case, in most of the configuration of parameters, the best results were obtained when the fully connected network was used.

%
\begin{table*}
\centering
\caption{Performance of our best methods evaluated using instances that were labeled with just one sense. Best results are marked in bold. Note that the proposed CN approach outperforms traditional approaches when using both F1 and Fuzzy NMI criteria. The results for the SemCor Ranked are not shown because, in the analysis considered only one possible sense, SemCor Ranked and SemCor MFS are equivalent.}
\begin{tabular}{@{\extracolsep{4pt}}cccccc}
\hline
System & $\omega$ & $k$ & F1 & Fuzzy NMI & Fuzzy B-Cubed \\
\hline
CN-ADD & 4 & - & \textbf{0.592} & 0.048 & 0.426 \\
CN-ADD & 2 & - & 0.554 & \textbf{0.049} & 0.356 \\
CN-ADD/AVG & 4 & 1 & 0.569 & 0.031 & \textbf{0.453} \\

\hline
One sense & - & - & 0.569 & 0.0 & 0.570 \\
1c1inst & - & - & 0.0 & 0.018 & 0.0 \\
SemCor MFS & - & - & 0.477 & 0.0 & 0.570 \\
\hline
\end{tabular}

\label{tab2}
\end{table*}

In the last assessment, only instances labeled with multiple senses were considered in the analysis. The obtained results are shown in Table \ref{tab3}. Considering the criterium based on ranking senses and quantifying their applicability, our method have had only results close to the participants and below the best baselines. However, our methods outperformed all participants in the detection of which senses are applicable (see Jaccard Index) and in both cluster evaluation criteria. Once again, most of the best results were obtained for a fully connected network in the $k$-NN connectivity method.

\begin{table*}[h]
\centering
\caption{Performance of our best methods evaluated using instances that were labeled with multiple senses. Best results are marked in bold.}
\begin{tabular}{@{\extracolsep{4pt}}cccccccc}
\hline
 &  &  & \multicolumn{3}{c}{ WSD F1} & \multicolumn{2}{c}{Cluster Comparison} \\
\cline{4-6} \cline{7-8}
System & $\omega$ & $k$ & Jaccard & $K_\delta^{sim}$ & WNDCG & Fuzzy NMI & Fuzzy B-Cubed \\
\hline
CN-ADD/AVG & 4 & 5 & \textbf{0.473} & 0.564 & 0.258 & 0.018 & 0.126 \\
CN-ADD/AVG & 7 & 1 & 0.438 & \textbf{0.604} & 0.257 & \textbf{0.040} & 0.131 \\
CN-ADD/AVG & 10 & - & 0.464 & 0.562 & \textbf{0.263} & 0.021 & \textbf{0.137} \\
CN-ADD/AVG & 4 & 1 & 0.441 & 0.595 & 0.256 & \textbf{0.040} & 0.129 \\

\hline
One sense & - & - & 0.387 & 0.635 & 0.254 & 0.0 & 0.130 \\
1c1inst & - & - & 0.0 & 0.0 & 0.0 & 0.300 & 0.0 \\
SemCor MFS & - & - & 0.283 & 0.373 & 0.197 & - & - \\
SemCor Ranked & - & - & 0.263 & 0.593 & 0.395 & - & - \\

\hline
\end{tabular}

\label{tab3}
\end{table*}

Overall, the proposed CN-based approach displayed competitive results in the the considered scenarios, either compared to baselines or compared to the participating systems. The use of addition and averaging to generate context embeddings turned out to be equivalent in many of the best obtained results, when considering the same parameters. It is also evident from the results that the performance of the proposed method varies with the type of ambiguity being tackled (single sense vs. multiple sense).
Concerning the variation in creating the embedding networks, it is worth mentioning that the \textit{fully-connected} model displayed the best performance in most of the cases. However, in some cases the $k$-NN model also displayed good results for particular values of $k$. Concerning the definition of the context window size, no clear pattern could be observed in Tables \ref{tab1}--\ref{tab3}. This means that the context size might depend on either the corpus some property related to the specificities of the ambiguous word. A further analysis of how the method depends on the parameters is provided in the next section.

\subsection{Parameter dependence}\label{sec:par}

\begin{figure}[ht]
	   \centering
       \includegraphics[width=0.95\textwidth]{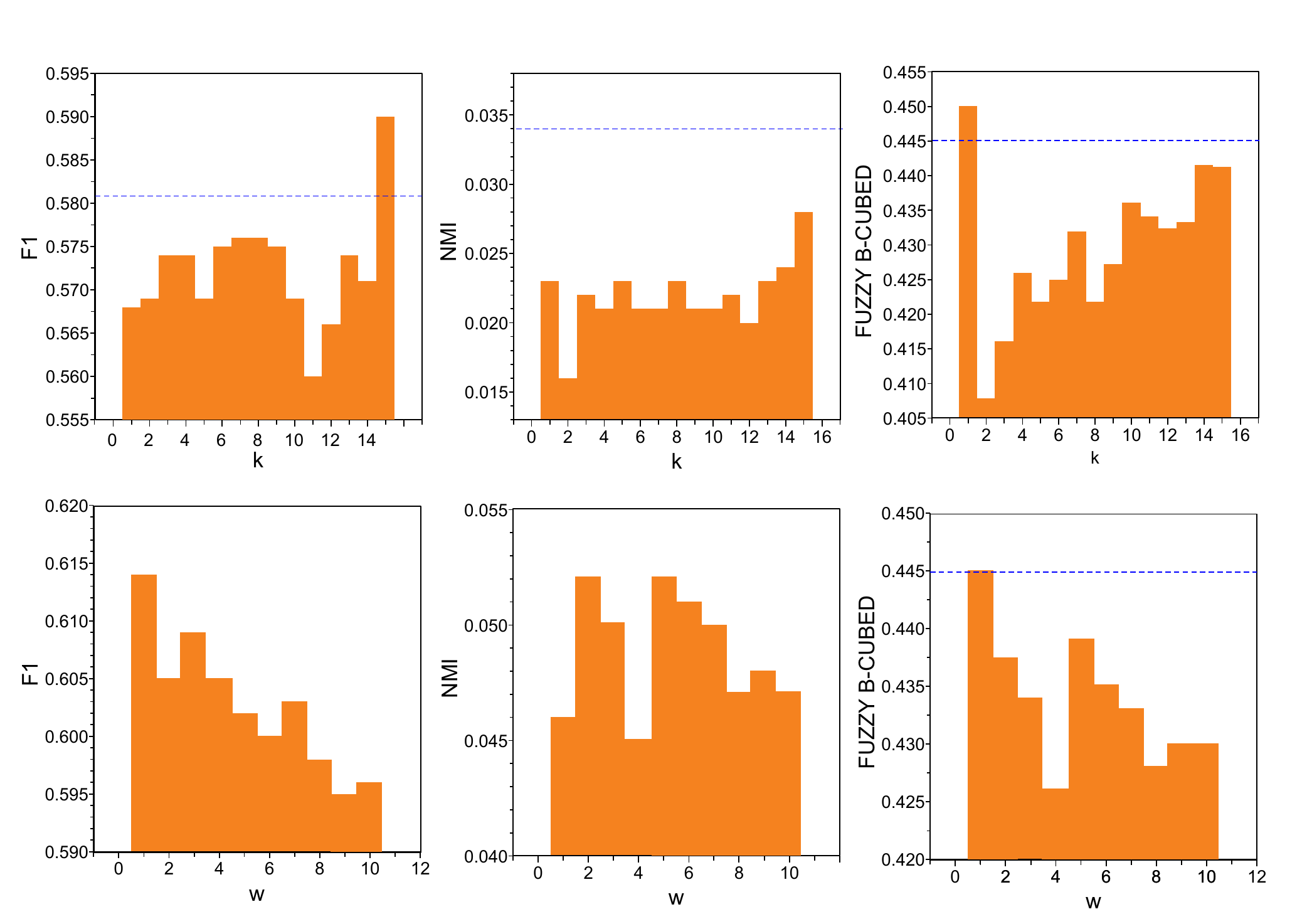}
       \caption{Dependence of the performance results using different configuration of parameters. In all figures, we show the scenario allowing only one sense for the ambiguous word. In (a), (b) and (c); we analyze the behavior of the performance as a function of $k$ when the full context of the sentence is considered. In (d), (e) and (f); we analyze the behavior of the performance for distinct values of $\omega$ when the fully-connected network is considered. The dashed lines represent the performance obtained with the full context (a-c) and the fully-connected network (f).}
       \label{fig2}
\end{figure}

In this section, we investigate the dependency of the results obtained by our method with the choice of parameters used to create the network. In Figure \ref{fig2}, we show the results obtained considered three criteria: F1, NMI and Fuzzy B-Cubed. Subfigures (a)-(c) analyze the performance obtained for different values of $k$, while subfigure (d)-(f) show the performance obtained when varying the context size $\omega$. The dashed lines represent the performance obtained when the \emph{fully-connected} strategy is used ((a)-(c)) or the full context of the sentence is used ((d)-(f)). No dashed lines are shown in (d) and (e) because the performance  obtained with the full context is much lower than the performance values shown for different values of $\omega$.

The variability of the performance with $k$ reveals that, in general, a good performance can be obtained with high values of $k$. In (a), (b) and (c), excellent performances were obtained for $k=15$. The fully-connected model also displayed an excellent performance in all three cases, being the best choice for the NMI index. These results confirm that the informativeness of the proposed model relies on both weak and strong ties, since optimized results are obtained mostly when all weighted links are considered.
We should note, however, that in particular cases the best performance is achieved with a single neighbor connection (see Figure \ref{fig2}(c)). Similar results can be observed for the other performance indexes, as shown in the \textcolor{black}{Supplementary Information}.

While the performance tends to be increased with high values of $k$, the best performance when $\omega$ varies seems to arise for the lowest values of context window. In (d) and (f), the optimum performance is obtained for $\omega=1$. In (e), the NMI is optimized when $\omega=2$. The full context only displays a good performance for the Fuzzy B-cubed measurement. Similar results were observed for the other measurements (\textcolor{black}{see the Supplementary Information}). Overall, the results showed that a low value of context is enough to provide good performance for the proposed model, considering both WSD-F1 and cluster comparison scenarios.

\section{Conclusion}\label{5}

In this paper, we explored the concept of \emph{context embeddings} modeled as complex networks to induce word senses via community detection algorithms. We evaluated multiple settings of our model and compared with well-known baselines and other systems that participated of the SemEval-2013 Task 13. We have shown that the proposed model presents a significant performance in both single and multiple senses multiple scenarios, without the use of annotated corpora, in a completely unsupervised manner. Moreover, we have shown that a good performance can be obtained when considering only a small context window to generate the embeddings. In a similar fashion, we have also found that, in general, a fully-connected and weighted network provides a better representation for the task.  The absence of any annotation allows the use of the proposed method in a range of graph-based applications in scenarios where unsupervised methods are required to process natural languages.

As future works, we intend to explore the use of community detection algorithms that provide soft communities instead of the hard communities provided by most of the current methods. We also intend to explore the use of neural language models to generate context embeddings in order to improve the quality of the context representation. Finally, we intend to integrate our methods with other natural language processing tasks~\cite{comparingStyle,10.1371/journal.pone.0136076,Ban171217,10.1371/journal.pone.0192545,10.1371/journal.pone.0187164,VIANA2013371} that might benefit from representing words as \emph{context embeddings}.

\section*{Acknowledgements}
\noindent
E.A.C.Jr. acknowledges financial support from Google (Research Awards in Latin America grant) and CAPES-Brazil. D.R.A. acknowledges financial support from Google (Research Awards in Latin America grant) and S\~ao Paulo Research Foundation (FAPESP) (grant no. 2014/20830-0, 2016/19069-9 and 2017/13464-6).



\begin{thebibliography}{54}
\expandafter\ifx\csname natexlab\endcsname\relax\def\natexlab#1{#1}\fi
\providecommand{\url}[1]{\texttt{#1}}
\providecommand{\href}[2]{#2}
\providecommand{\path}[1]{#1}
\providecommand{\DOIprefix}{doi:}
\providecommand{\ArXivprefix}{arXiv:}
\providecommand{\URLprefix}{URL: }
\providecommand{\Pubmedprefix}{pmid:}
\providecommand{\doi}[1]{\href{http://dx.doi.org/#1}{\path{#1}}}
\providecommand{\Pubmed}[1]{\href{pmid:#1}{\path{#1}}}
\providecommand{\bibinfo}[2]{#2}
\ifx\xfnm\relax \def\xfnm[#1]{\unskip,\space#1}\fi
\bibitem[{Agirre et~al.(2009)Agirre, De~Lacalle \&
  Soroa}]{Agirre:2009:KWS:1661445.1661686}
\bibinfo{author}{Agirre, E.}, \bibinfo{author}{De~Lacalle, O.~L.}, \&
  \bibinfo{author}{Soroa, A.} (\bibinfo{year}{2009}).
\newblock \bibinfo{title}{Knowledge-based wsd on specific domains: Performing
  better than generic supervised wsd}.
\newblock In {\it \bibinfo{booktitle}{Proceedings of the 21st International
  Jont Conference on Artifical Intelligence}\/} (pp.
  \bibinfo{pages}{1501--1506}).
\newblock \bibinfo{address}{San Francisco, CA, USA}: \bibinfo{publisher}{Morgan
  Kaufmann Publishers Inc.}
\bibitem[{Agirre \& Soroa(2007)}]{agirre2007semeval}
\bibinfo{author}{Agirre, E.}, \& \bibinfo{author}{Soroa, A.}
  (\bibinfo{year}{2007}).
\newblock \bibinfo{title}{Semeval-2007 task 02: Evaluating word sense induction
  and discrimination systems}.
\newblock In {\it \bibinfo{booktitle}{Proceedings of the 4th International
  Workshop on Semantic Evaluations}\/} (pp. \bibinfo{pages}{7--12}).
\newblock \bibinfo{organization}{Association for Computational Linguistics}.
\bibitem[{Agirre \& Soroa(2009)}]{agirre2009personalizing}
\bibinfo{author}{Agirre, E.}, \& \bibinfo{author}{Soroa, A.}
  (\bibinfo{year}{2009}).
\newblock \bibinfo{title}{Personalizing pagerank for word sense
  disambiguation}.
\newblock In {\it \bibinfo{booktitle}{Proceedings of the 12th Conference of the
  European Chapter of the Association for Computational Linguistics}\/} (pp.
  \bibinfo{pages}{33--41}).
\newblock \bibinfo{organization}{Association for Computational Linguistics}.
\bibitem[{Amancio(2015{\natexlab{a}})}]{comparingStyle}
\bibinfo{author}{Amancio, D.~R.} (\bibinfo{year}{2015}{\natexlab{a}}).
\newblock \bibinfo{title}{Comparing the topological properties of real and
  artificially generated scientific manuscripts}.
\newblock {\it \bibinfo{journal}{Scientometrics}\/},  {\it
  \bibinfo{volume}{105}\/}, \bibinfo{pages}{1763--1779}.
\bibitem[{Amancio(2015{\natexlab{b}})}]{10.1371/journal.pone.0136076}
\bibinfo{author}{Amancio, D.~R.} (\bibinfo{year}{2015}{\natexlab{b}}).
\newblock \bibinfo{title}{A complex network approach to stylometry}.
\newblock {\it \bibinfo{journal}{PLoS ONE}\/},  {\it \bibinfo{volume}{10}\/},
  \bibinfo{pages}{e0136076}.
\bibitem[{Amancio et~al.(2012)Amancio, Oliveira~Jr. \&
  Costa}]{0295-5075-98-1-18002}
\bibinfo{author}{Amancio, D.~R.}, \bibinfo{author}{Oliveira~Jr., O.~N.}, \&
  \bibinfo{author}{Costa, L.~F.} (\bibinfo{year}{2012}).
\newblock \bibinfo{title}{Unveiling the relationship between complex networks
  metrics and word senses}.
\newblock {\it \bibinfo{journal}{EPL (Europhysics Letters)}\/},  {\it
  \bibinfo{volume}{98}\/}, \bibinfo{pages}{18002}.
\bibitem[{Ban et~al.(2017)Ban, Perc \& Levnaji{\'c}}]{Ban171217}
\bibinfo{author}{Ban, K.}, \bibinfo{author}{Perc, M.}, \&
  \bibinfo{author}{Levnaji{\'c}, Z.} (\bibinfo{year}{2017}).
\newblock \bibinfo{title}{Robust clustering of languages across wikipedia
  growth}.
\newblock {\it \bibinfo{journal}{Royal Society Open Science}\/},  {\it
  \bibinfo{volume}{4}\/}.
\bibitem[{Baroni et~al.(2014)Baroni, Dinu \& Kruszewski}]{baroni2014don}
\bibinfo{author}{Baroni, M.}, \bibinfo{author}{Dinu, G.}, \&
  \bibinfo{author}{Kruszewski, G.} (\bibinfo{year}{2014}).
\newblock \bibinfo{title}{Don't count, predict! a systematic comparison of
  context-counting vs. context-predicting semantic vectors.}
\newblock In {\it \bibinfo{booktitle}{ACL (1)}\/} (pp.
  \bibinfo{pages}{238--247}).
\bibitem[{Baskaya et~al.(2013)Baskaya, Sert, Cirik \& Yuret}]{baskaya2013ai}
\bibinfo{author}{Baskaya, O.}, \bibinfo{author}{Sert, E.},
  \bibinfo{author}{Cirik, V.}, \& \bibinfo{author}{Yuret, D.}
  (\bibinfo{year}{2013}).
\newblock \bibinfo{title}{Ai-ku: Using substitute vectors and co-occurrence
  modeling for word sense induction and disambiguation}.
\newblock In {\it \bibinfo{booktitle}{Second Joint Conference on Lexical and
  Computational Semantics (* SEM), Volume 2: Proceedings of the Seventh
  International Workshop on Semantic Evaluation (SemEval 2013)}\/} (pp.
  \bibinfo{pages}{300--306}).
\newblock volume~\bibinfo{volume}{2}.
\bibitem[{Bengio et~al.(2003)Bengio, Ducharme, Vincent \&
  Jauvin}]{bengio2003neural}
\bibinfo{author}{Bengio, Y.}, \bibinfo{author}{Ducharme, R.},
  \bibinfo{author}{Vincent, P.}, \& \bibinfo{author}{Jauvin, C.}
  (\bibinfo{year}{2003}).
\newblock \bibinfo{title}{A neural probabilistic language model}.
\newblock {\it \bibinfo{journal}{journal of machine learning research}\/},
  {\it \bibinfo{volume}{3}\/}, \bibinfo{pages}{1137--1155}.
\bibitem[{Blondel et~al.(2008)Blondel, Guillaume, Lambiotte \&
  Lefebvre}]{blondel2008fast}
\bibinfo{author}{Blondel, V.~D.}, \bibinfo{author}{Guillaume, J.-L.},
  \bibinfo{author}{Lambiotte, R.}, \& \bibinfo{author}{Lefebvre, E.}
  (\bibinfo{year}{2008}).
\newblock \bibinfo{title}{Fast unfolding of communities in large networks}.
\newblock {\it \bibinfo{journal}{Journal of statistical mechanics: theory and
  experiment}\/},  {\it \bibinfo{volume}{2008}\/}, \bibinfo{pages}{P10008}.
\bibitem[{Breve \& Zhao(2013)}]{Breve2013}
\bibinfo{author}{Breve, F.}, \& \bibinfo{author}{Zhao, L.}
  (\bibinfo{year}{2013}).
\newblock \bibinfo{title}{Fuzzy community structure detection by particle
  competition and cooperation}.
\newblock {\it \bibinfo{journal}{Soft Computing}\/},  {\it
  \bibinfo{volume}{17}\/}, \bibinfo{pages}{659--673}.
\bibitem[{Breve et~al.(2012)Breve, Zhao, Quiles, Pedrycz \& Liu}]{5871621}
\bibinfo{author}{Breve, F.}, \bibinfo{author}{Zhao, L.},
  \bibinfo{author}{Quiles, M.}, \bibinfo{author}{Pedrycz, W.}, \&
  \bibinfo{author}{Liu, J.} (\bibinfo{year}{2012}).
\newblock \bibinfo{title}{Particle competition and cooperation in networks for
  semi-supervised learning}.
\newblock {\it \bibinfo{journal}{IEEE Transactions on Knowledge and Data
  Engineering}\/},  {\it \bibinfo{volume}{24}\/}, \bibinfo{pages}{1686--1698}.
\bibitem[{Chen et~al.(2018)Chen, Chen \& Liu}]{10.1371/journal.pone.0192545}
\bibinfo{author}{Chen, H.}, \bibinfo{author}{Chen, X.}, \&
  \bibinfo{author}{Liu, H.} (\bibinfo{year}{2018}).
\newblock \bibinfo{title}{How does language change as a lexical network? an
  investigation based on written chinese word co-occurrence networks}.
\newblock {\it \bibinfo{journal}{PLoS ONE}\/},  {\it \bibinfo{volume}{13}\/},
  \bibinfo{pages}{e0187164}.
\bibitem[{Collobert \& Weston(2008)}]{collobert2008unified}
\bibinfo{author}{Collobert, R.}, \& \bibinfo{author}{Weston, J.}
  (\bibinfo{year}{2008}).
\newblock \bibinfo{title}{A unified architecture for natural language
  processing: Deep neural networks with multitask learning}.
\newblock In {\it \bibinfo{booktitle}{Proceedings of the 25th international
  conference on Machine learning}\/} (pp. \bibinfo{pages}{160--167}).
\newblock \bibinfo{organization}{ACM}.
\bibitem[{Collobert et~al.(2011)Collobert, Weston, Bottou, Karlen, Kavukcuoglu
  \& Kuksa}]{collobert2011natural}
\bibinfo{author}{Collobert, R.}, \bibinfo{author}{Weston, J.},
  \bibinfo{author}{Bottou, L.}, \bibinfo{author}{Karlen, M.},
  \bibinfo{author}{Kavukcuoglu, K.}, \& \bibinfo{author}{Kuksa, P.}
  (\bibinfo{year}{2011}).
\newblock \bibinfo{title}{Natural language processing (almost) from scratch}.
\newblock {\it \bibinfo{journal}{Journal of Machine Learning Research}\/},
  {\it \bibinfo{volume}{12}\/}, \bibinfo{pages}{2493--2537}.
\bibitem[{Corr\^ea~Jr. et~al.(2018)Corr\^ea~Jr., Lopes \&
  Amancio}]{CORREA2018103}
\bibinfo{author}{Corr\^ea~Jr., E.}, \bibinfo{author}{Lopes, A.~A.}, \&
  \bibinfo{author}{Amancio, D.~R.} (\bibinfo{year}{2018}).
\newblock \bibinfo{title}{Word sense disambiguation: A complex network
  approach}.
\newblock {\it \bibinfo{journal}{Information Sciences}\/},  {\it
  \bibinfo{volume}{442-443}\/}, \bibinfo{pages}{103 -- 113}.
\bibitem[{Fortunato(2010)}]{FORTUNATO201075}
\bibinfo{author}{Fortunato, S.} (\bibinfo{year}{2010}).
\newblock \bibinfo{title}{Community detection in graphs}.
\newblock {\it \bibinfo{journal}{Physics Reports}\/},  {\it
  \bibinfo{volume}{486}\/}, \bibinfo{pages}{75 -- 174}.
\bibitem[{Gale et~al.(1992)Gale, Church \& Yarowsky}]{gale1992method}
\bibinfo{author}{Gale, W.~A.}, \bibinfo{author}{Church, K.~W.}, \&
  \bibinfo{author}{Yarowsky, D.} (\bibinfo{year}{1992}).
\newblock \bibinfo{title}{A method for disambiguating word senses in a large
  corpus}.
\newblock {\it \bibinfo{journal}{Computers and the Humanities}\/},  {\it
  \bibinfo{volume}{26}\/}, \bibinfo{pages}{415--439}.
\bibitem[{Gao et~al.(2013)Gao, Zhang, Jin, Marwan \&
  Kurths}]{PhysRevE.88.032910}
\bibinfo{author}{Gao, Z.-K.}, \bibinfo{author}{Zhang, X.-W.},
  \bibinfo{author}{Jin, N.-D.}, \bibinfo{author}{Marwan, N.}, \&
  \bibinfo{author}{Kurths, J.} (\bibinfo{year}{2013}).
\newblock \bibinfo{title}{Multivariate recurrence network analysis for
  characterizing horizontal oil-water two-phase flow}.
\newblock {\it \bibinfo{journal}{Phys. Rev. E}\/},  {\it
  \bibinfo{volume}{88}\/}, \bibinfo{pages}{032910}.
\bibitem[{Goyal \& Hovy(2014)}]{goyal2014unsupervised}
\bibinfo{author}{Goyal, K.}, \& \bibinfo{author}{Hovy, E.~H.}
  (\bibinfo{year}{2014}).
\newblock \bibinfo{title}{Unsupervised word sense induction using
  distributional statistics.}
\newblock In {\it \bibinfo{booktitle}{COLING}\/} (pp.
  \bibinfo{pages}{1302--1310}).
\bibitem[{Hope \& Keller(2013)}]{hope2013uos}
\bibinfo{author}{Hope, D.}, \& \bibinfo{author}{Keller, B.}
  (\bibinfo{year}{2013}).
\newblock \bibinfo{title}{Uos: A graph-based system for graded word sense
  induction}.
\newblock In {\it \bibinfo{booktitle}{Second Joint Conference on Lexical and
  Computational Semantics (* SEM), Volume 2: Proceedings of the Seventh
  International Workshop on Semantic Evaluation (SemEval 2013)}\/} (pp.
  \bibinfo{pages}{689--694}).
\newblock volume~\bibinfo{volume}{2}.
\bibitem[{Iacobacci et~al.(2015)Iacobacci, Pilehvar \&
  Navigli}]{iacobacci2015sensembed}
\bibinfo{author}{Iacobacci, I.}, \bibinfo{author}{Pilehvar, M.~T.}, \&
  \bibinfo{author}{Navigli, R.} (\bibinfo{year}{2015}).
\newblock \bibinfo{title}{Sensembed: learning sense embeddings for word and
  relational similarity}.
\newblock In {\it \bibinfo{booktitle}{Proceedings of ACL}\/} (pp.
  \bibinfo{pages}{95--105}).
\bibitem[{Iacobacci et~al.(2016)Iacobacci, Pilehvar \&
  Navigli}]{iacobacci2016embeddings}
\bibinfo{author}{Iacobacci, I.}, \bibinfo{author}{Pilehvar, M.~T.}, \&
  \bibinfo{author}{Navigli, R.} (\bibinfo{year}{2016}).
\newblock \bibinfo{title}{Embeddings for word sense disambiguation: An
  evaluation study}.
\newblock In {\it \bibinfo{booktitle}{Proceedings of the 54th Annual Meeting of
  the Association for Computational Linguistics (Volume 1: Long Papers)}\/}
  (pp. \bibinfo{pages}{897--907}).
\newblock \bibinfo{address}{Berlin, Germany}: \bibinfo{publisher}{Association
  for Computational Linguistics}.
\bibitem[{Jurgens \& Klapaftis(2013)}]{jurgens2013semeval}
\bibinfo{author}{Jurgens, D.}, \& \bibinfo{author}{Klapaftis, I.}
  (\bibinfo{year}{2013}).
\newblock \bibinfo{title}{Semeval-2013 task 13: Word sense induction for graded
  and non-graded senses}.
\newblock In {\it \bibinfo{booktitle}{Second joint conference on lexical and
  computational semantics (* SEM)}\/} (pp. \bibinfo{pages}{290--299}).
\newblock volume~\bibinfo{volume}{2}.
\bibitem[{K{\aa}geb{\"a}ck et~al.(2015)K{\aa}geb{\"a}ck, Johansson, Johansson
  \& Dubhashi}]{kaageback2015neural}
\bibinfo{author}{K{\aa}geb{\"a}ck, M.}, \bibinfo{author}{Johansson, F.},
  \bibinfo{author}{Johansson, R.}, \& \bibinfo{author}{Dubhashi, D.}
  (\bibinfo{year}{2015}).
\newblock \bibinfo{title}{Neural context embeddings for automatic discovery of
  word senses}.
\newblock In {\it \bibinfo{booktitle}{Proceedings of NAACL-HLT}\/} (pp.
  \bibinfo{pages}{25--32}).
\bibitem[{Lau et~al.(2013)Lau, Cook \& Baldwin}]{lau2013unimelb}
\bibinfo{author}{Lau, J.~H.}, \bibinfo{author}{Cook, P.}, \&
  \bibinfo{author}{Baldwin, T.} (\bibinfo{year}{2013}).
\newblock \bibinfo{title}{unimelb: Topic modelling-based word sense induction}.
\newblock In {\it \bibinfo{booktitle}{Second Joint Conference on Lexical and
  Computational Semantics (* SEM), Volume 2: Proceedings of the Seventh
  International Workshop on Semantic Evaluation (SemEval 2013)}\/} (pp.
  \bibinfo{pages}{307--311}).
\newblock volume~\bibinfo{volume}{2}.
\bibitem[{Lin(1998)}]{Lin}
\bibinfo{author}{Lin, D.} (\bibinfo{year}{1998}).
\newblock \bibinfo{title}{Automatic retrieval and clustering of similar words}.
\newblock In {\it \bibinfo{booktitle}{Proceedings of the 17th International
  Conference on Computational Linguistics - Volume 2}\/} COLING '98 (pp.
  \bibinfo{pages}{768--774}).
\newblock \bibinfo{address}{Stroudsburg, PA, USA}:
  \bibinfo{publisher}{Association for Computational Linguistics}.
\bibitem[{Liu et~al.(2015)Liu, Liu, Chua \& Sun}]{Liu:2015:TWE:2886521.2886657}
\bibinfo{author}{Liu, Y.}, \bibinfo{author}{Liu, Z.}, \bibinfo{author}{Chua,
  T.-S.}, \& \bibinfo{author}{Sun, M.} (\bibinfo{year}{2015}).
\newblock \bibinfo{title}{Topical word embeddings}.
\newblock In {\it \bibinfo{booktitle}{Proceedings of the Twenty-Ninth AAAI
  Conference on Artificial Intelligence}\/} AAAI'15 (pp.
  \bibinfo{pages}{2418--2424}).
\newblock \bibinfo{publisher}{AAAI Press}.
\bibitem[{Manandhar et~al.(2010)Manandhar, Klapaftis, Dligach \&
  Pradhan}]{manandhar2010semeval}
\bibinfo{author}{Manandhar, S.}, \bibinfo{author}{Klapaftis, I.~P.},
  \bibinfo{author}{Dligach, D.}, \& \bibinfo{author}{Pradhan, S.~S.}
  (\bibinfo{year}{2010}).
\newblock \bibinfo{title}{Semeval-2010 task 14: Word sense induction \&
  disambiguation}.
\newblock In {\it \bibinfo{booktitle}{Proceedings of the 5th international
  workshop on semantic evaluation}\/} (pp. \bibinfo{pages}{63--68}).
\newblock \bibinfo{organization}{Association for Computational Linguistics}.
\bibitem[{Manning et~al.(2008)Manning, Raghavan \&
  Sch\"{u}tze}]{Manning:2008:IIR:1394399}
\bibinfo{author}{Manning, C.~D.}, \bibinfo{author}{Raghavan, P.}, \&
  \bibinfo{author}{Sch\"{u}tze, H.} (\bibinfo{year}{2008}).
\newblock {\it \bibinfo{title}{Introduction to Information Retrieval}\/}.
\newblock \bibinfo{address}{New York, NY, USA}: \bibinfo{publisher}{Cambridge
  University Press}.
\bibitem[{Mikolov et~al.(2013{\natexlab{a}})Mikolov, Chen, Corrado \&
  Dean}]{mikolov2013efficient}
\bibinfo{author}{Mikolov, T.}, \bibinfo{author}{Chen, K.},
  \bibinfo{author}{Corrado, G.}, \& \bibinfo{author}{Dean, J.}
  (\bibinfo{year}{2013}{\natexlab{a}}).
\newblock \bibinfo{title}{Efficient estimation of word representations in
  vector space}.
\newblock {\it \bibinfo{journal}{arXiv preprint arXiv:1301.3781}\/}, .
\bibitem[{Mikolov et~al.(2013{\natexlab{b}})Mikolov, Sutskever, Chen, Corrado
  \& Dean}]{mikolov2013distributed}
\bibinfo{author}{Mikolov, T.}, \bibinfo{author}{Sutskever, I.},
  \bibinfo{author}{Chen, K.}, \bibinfo{author}{Corrado, G.~S.}, \&
  \bibinfo{author}{Dean, J.} (\bibinfo{year}{2013}{\natexlab{b}}).
\newblock \bibinfo{title}{Distributed representations of words and phrases and
  their compositionality}.
\newblock In {\it \bibinfo{booktitle}{Advances in neural information processing
  systems}\/} (pp. \bibinfo{pages}{3111--3119}).
\bibitem[{Mnih \& Kavukcuoglu(2013)}]{NIPS2013_5165}
\bibinfo{author}{Mnih, A.}, \& \bibinfo{author}{Kavukcuoglu, K.}
  (\bibinfo{year}{2013}).
\newblock \bibinfo{title}{Learning word embeddings efficiently with
  noise-contrastive estimation}.
\newblock In \bibinfo{editor}{C.~J.~C. Burges}, \bibinfo{editor}{L.~Bottou},
  \bibinfo{editor}{M.~Welling}, \bibinfo{editor}{Z.~Ghahramani}, \&
  \bibinfo{editor}{K.~Q. Weinberger} (Eds.), {\it \bibinfo{booktitle}{Advances
  in Neural Information Processing Systems 26}\/} (pp.
  \bibinfo{pages}{2265--2273}).
\newblock \bibinfo{publisher}{Curran Associates, Inc.}
\bibitem[{Navigli(2009)}]{navigli2009word}
\bibinfo{author}{Navigli, R.} (\bibinfo{year}{2009}).
\newblock \bibinfo{title}{Word sense disambiguation: A survey}.
\newblock {\it \bibinfo{journal}{ACM Computing Surveys (CSUR)}\/},  {\it
  \bibinfo{volume}{41}\/}, \bibinfo{pages}{10}.
\bibitem[{Navigli \& Vannella(2013)}]{navigli2013semeval}
\bibinfo{author}{Navigli, R.}, \& \bibinfo{author}{Vannella, D.}
  (\bibinfo{year}{2013}).
\newblock \bibinfo{title}{Semeval-2013 task 11: Word sense induction and
  disambiguation within an end-user application}.
\newblock In {\it \bibinfo{booktitle}{Second Joint Conference on Lexical and
  Computational Semantics (* SEM)}\/} (pp. \bibinfo{pages}{193--201}).
\newblock volume~\bibinfo{volume}{2}.
\bibitem[{Newman(2006)}]{newman2006modularity}
\bibinfo{author}{Newman, M.~E.} (\bibinfo{year}{2006}).
\newblock \bibinfo{title}{Modularity and community structure in networks}.
\newblock {\it \bibinfo{journal}{Proceedings of the national academy of
  sciences}\/},  {\it \bibinfo{volume}{103}\/}, \bibinfo{pages}{8577--8582}.
\bibitem[{Pennington et~al.(2014)Pennington, Socher \&
  Manning}]{pennington2014glove}
\bibinfo{author}{Pennington, J.}, \bibinfo{author}{Socher, R.}, \&
  \bibinfo{author}{Manning, C.~D.} (\bibinfo{year}{2014}).
\newblock \bibinfo{title}{Glove: Global vectors for word representation}.
\newblock In {\it \bibinfo{booktitle}{Empirical Methods in Natural Language
  Processing (EMNLP)}\/} (pp. \bibinfo{pages}{1532--1543}).
\bibitem[{Perozzi et~al.(2014)Perozzi, Al-Rfou, Kulkarni \&
  Skiena}]{perozzi2014inducing}
\bibinfo{author}{Perozzi, B.}, \bibinfo{author}{Al-Rfou, R.},
  \bibinfo{author}{Kulkarni, V.}, \& \bibinfo{author}{Skiena, S.}
  (\bibinfo{year}{2014}).
\newblock \bibinfo{title}{Inducing language networks from continuous space word
  representations}.
\newblock In {\it \bibinfo{booktitle}{Complex Networks V}\/} (pp.
  \bibinfo{pages}{261--273}).
\newblock \bibinfo{publisher}{Springer}.
\bibitem[{Rodriguez et~al.(2016)Rodriguez, Comin, Casanova, Bruno, Amancio,
  Rodrigues \& Costa}]{comparacoes}
\bibinfo{author}{Rodriguez, M.~Z.}, \bibinfo{author}{Comin, C.~H.},
  \bibinfo{author}{Casanova, D.}, \bibinfo{author}{Bruno, O.~M.},
  \bibinfo{author}{Amancio, D.~R.}, \bibinfo{author}{Rodrigues, F.~A.}, \&
  \bibinfo{author}{Costa, L.~F.} (\bibinfo{year}{2016}).
\newblock \bibinfo{title}{Clustering algorithms: a comparative approach}.
\newblock {\it \bibinfo{journal}{arXiv:1612.08388}\/}, .
\bibitem[{Sagae \& Gordon(2009)}]{Sagae}
\bibinfo{author}{Sagae, K.}, \& \bibinfo{author}{Gordon, A.~S.}
  (\bibinfo{year}{2009}).
\newblock \bibinfo{title}{Clustering words by syntactic similarity improves
  dependency parsing of predicate-argument structures}.
\newblock In {\it \bibinfo{booktitle}{Proceedings of the 11th International
  Conference on Parsing Technologies}\/} IWPT 09 (pp.
  \bibinfo{pages}{192--201}).
\newblock \bibinfo{address}{Stroudsburg, PA, USA}:
  \bibinfo{publisher}{Association for Computational Linguistics}.
\bibitem[{Schnabel et~al.(2015)Schnabel, Labutov, Mimno \&
  Joachims}]{schnabel-EtAl:2015:EMNLP}
\bibinfo{author}{Schnabel, T.}, \bibinfo{author}{Labutov, I.},
  \bibinfo{author}{Mimno, D.}, \& \bibinfo{author}{Joachims, T.}
  (\bibinfo{year}{2015}).
\newblock \bibinfo{title}{Evaluation methods for unsupervised word embeddings}.
\newblock In {\it \bibinfo{booktitle}{Proceedings of the 2015 Conference on
  Empirical Methods in Natural Language Processing}\/} (pp.
  \bibinfo{pages}{298--307}).
\newblock \bibinfo{address}{Lisbon, Portugal}: \bibinfo{publisher}{Association
  for Computational Linguistics}.
\bibitem[{Silva et~al.(2016)Silva, Amancio, Bardosova, Costa \&
  Oliveira}]{silva2016using}
\bibinfo{author}{Silva, F.~N.}, \bibinfo{author}{Amancio, D.~R.},
  \bibinfo{author}{Bardosova, M.}, \bibinfo{author}{Costa, L. d.~F.}, \&
  \bibinfo{author}{Oliveira, O.~N.} (\bibinfo{year}{2016}).
\newblock \bibinfo{title}{Using network science and text analytics to produce
  surveys in a scientific topic}.
\newblock {\it \bibinfo{journal}{Journal of Informetrics}\/},  {\it
  \bibinfo{volume}{10}\/}, \bibinfo{pages}{487--502}.
\bibitem[{Silva \& Amancio(2012)}]{0295-5075-98-5-58001}
\bibinfo{author}{Silva, T.~C.}, \& \bibinfo{author}{Amancio, D.~R.}
  (\bibinfo{year}{2012}).
\newblock \bibinfo{title}{Word sense disambiguation via high order of learning
  in complex networks}.
\newblock {\it \bibinfo{journal}{EPL (Europhysics Letters)}\/},  {\it
  \bibinfo{volume}{98}\/}, \bibinfo{pages}{58001}.
\bibitem[{Sugawara et~al.(2016)Sugawara, Takamura, Sasano \&
  Okumura}]{10.1007/978-981-10-0515-2_8}
\bibinfo{author}{Sugawara, H.}, \bibinfo{author}{Takamura, H.},
  \bibinfo{author}{Sasano, R.}, \& \bibinfo{author}{Okumura, M.}
  (\bibinfo{year}{2016}).
\newblock \bibinfo{title}{Context representation with word embeddings for wsd}.
\newblock In \bibinfo{editor}{K.~Hasida}, \& \bibinfo{editor}{A.~Purwarianti}
  (Eds.), {\it \bibinfo{booktitle}{Computational Linguistics}\/} (pp.
  \bibinfo{pages}{108--119}).
\newblock \bibinfo{address}{Singapore}: \bibinfo{publisher}{Springer
  Singapore}.
\bibitem[{Taghipour \& Ng(2015)}]{taghipour2015semi}
\bibinfo{author}{Taghipour, K.}, \& \bibinfo{author}{Ng, H.~T.}
  (\bibinfo{year}{2015}).
\newblock \bibinfo{title}{Semi-supervised word sense disambiguation using word
  embeddings in general and specific domains}.
\newblock In {\it \bibinfo{booktitle}{The 2015 Annual Conference of the North
  American Chapter of the Association for Computational Linguistics}\/} (pp.
  \bibinfo{pages}{314--323}).
\bibitem[{Uren et~al.(2006)Uren, Cimiano, Iria, Handschuh, Vargas-Vera, Motta
  \& Ciravegna}]{UREN200614}
\bibinfo{author}{Uren, V.}, \bibinfo{author}{Cimiano, P.},
  \bibinfo{author}{Iria, J.}, \bibinfo{author}{Handschuh, S.},
  \bibinfo{author}{Vargas-Vera, M.}, \bibinfo{author}{Motta, E.}, \&
  \bibinfo{author}{Ciravegna, F.} (\bibinfo{year}{2006}).
\newblock \bibinfo{title}{Semantic annotation for knowledge management:
  Requirements and a survey of the state of the art}.
\newblock {\it \bibinfo{journal}{Web Semantics: Science, Services and Agents on
  the World Wide Web}\/},  {\it \bibinfo{volume}{4}\/}, \bibinfo{pages}{14 --
  28}.
\bibitem[{V{\'e}ronis(2004)}]{veronis2004hyperlex}
\bibinfo{author}{V{\'e}ronis, J.} (\bibinfo{year}{2004}).
\newblock \bibinfo{title}{Hyperlex: lexical cartography for information
  retrieval}.
\newblock {\it \bibinfo{journal}{Computer Speech \& Language}\/},  {\it
  \bibinfo{volume}{18}\/}, \bibinfo{pages}{223--252}.
\bibitem[{Viana et~al.(2013)Viana, Amancio \& Costa}]{VIANA2013371}
\bibinfo{author}{Viana, M.~P.}, \bibinfo{author}{Amancio, D.~R.}, \&
  \bibinfo{author}{Costa, L.~F.} (\bibinfo{year}{2013}).
\newblock \bibinfo{title}{On time-varying collaboration networks}.
\newblock {\it \bibinfo{journal}{Journal of Informetrics}\/},  {\it
  \bibinfo{volume}{7}\/}, \bibinfo{pages}{371 -- 378}.
\bibitem[{Widdows \& Dorow(2002)}]{widdows2002graph}
\bibinfo{author}{Widdows, D.}, \& \bibinfo{author}{Dorow, B.}
  (\bibinfo{year}{2002}).
\newblock \bibinfo{title}{A graph model for unsupervised lexical acquisition}.
\newblock In {\it \bibinfo{booktitle}{Proceedings of the 19th international
  conference on Computational linguistics-Volume 1}\/} (pp.
  \bibinfo{pages}{1--7}).
\newblock \bibinfo{organization}{Association for Computational Linguistics}.
\bibitem[{Wilks \& Stevenson(1997)}]{W97-0208}
\bibinfo{author}{Wilks, Y.}, \& \bibinfo{author}{Stevenson, M.}
  (\bibinfo{year}{1997}).
\newblock \bibinfo{title}{Sense tagging: Semantic tagging with a lexicon}.
\newblock In {\it \bibinfo{booktitle}{Tagging Text with Lexical Semantics: Why,
  What, and How?}\/}.
\bibitem[{Yaveroğlu et~al.(2014)Yaveroğlu, Malod-Dognin, Davis, Levnajic,
  Janjic, Karapandza, Stojmirovic \& Pržulj}]{zoran}
\bibinfo{author}{Yaveroğlu, O.~N.}, \bibinfo{author}{Malod-Dognin, N.},
  \bibinfo{author}{Davis, D.}, \bibinfo{author}{Levnajic, Z.},
  \bibinfo{author}{Janjic, V.}, \bibinfo{author}{Karapandza, R.},
  \bibinfo{author}{Stojmirovic, A.}, \& \bibinfo{author}{Pržulj, N.}
  (\bibinfo{year}{2014}).
\newblock \bibinfo{title}{Revealing the hidden language of complex networks}.
\newblock {\it \bibinfo{journal}{Scientific Reports}\/},  {\it
  \bibinfo{volume}{4}\/}, \bibinfo{pages}{4547}.
\bibitem[{Yu et~al.(2017)Yu, Wang, Zhang, Zhang \&
  Liu}]{10.1371/journal.pone.0187164}
\bibinfo{author}{Yu, D.}, \bibinfo{author}{Wang, W.}, \bibinfo{author}{Zhang,
  S.}, \bibinfo{author}{Zhang, W.}, \& \bibinfo{author}{Liu, R.}
  (\bibinfo{year}{2017}).
\newblock \bibinfo{title}{Hybrid self-optimized clustering model based on
  citation links and textual features to detect research topics}.
\newblock {\it \bibinfo{journal}{PLoS ONE}\/},  {\it \bibinfo{volume}{12}\/},
  \bibinfo{pages}{e0187164}.
\bibitem[{Zhang et~al.(2014)Zhang, Liu, Li, Zhou \&
  Zong}]{zhang-EtAl:2014:P14-11}
\bibinfo{author}{Zhang, J.}, \bibinfo{author}{Liu, S.}, \bibinfo{author}{Li,
  M.}, \bibinfo{author}{Zhou, M.}, \& \bibinfo{author}{Zong, C.}
  (\bibinfo{year}{2014}).
\newblock \bibinfo{title}{Bilingually-constrained phrase embeddings for machine
  translation}.
\newblock In {\it \bibinfo{booktitle}{Proceedings of the 52nd Annual Meeting of
  the Association for Computational Linguistics (Volume 1: Long Papers)}\/}
  (pp. \bibinfo{pages}{111--121}).
\newblock \bibinfo{address}{Baltimore, Maryland}:
  \bibinfo{publisher}{Association for Computational Linguistics}.

\end{thebibliography}

\newpage

\section*{Supplementary Information}

The results below summarize the performance obtained with the proposed method by considering the variation of both context window ($\omega$) and network connectivity ($k$) in the $k$-NN model. The results are divided according to the following classification of studied instances: (a) all instances; (b) instances labeled with just one sense; and (c) instances labeled with multiple senses.

\begin{longtable}{cccccccc}
\caption{Performance of our best methods evaluated using all instances available in the shared task.}\\
\hline
 &  &  & \multicolumn{3}{c}{ WSD F1} & \multicolumn{2}{c}{Cluster Comparison} \\
\cline{4-6} \cline{7-8}
System & $\omega$ & $k$ & Jac. Ind. & $K_\delta^{sim}$ & WNDCG & Fuzzy NMI & Fuzzy B-Cubed \\
\hline
\endfirsthead

\caption{Performance of our best methods evaluated using all instances available in the shared task. (continuation)}\\
\hline
 &  &  & \multicolumn{3}{c}{WSD F1} & \multicolumn{2}{c}{Cluster Comparison} \\
\cline{4-6} \cline{7-8}
System & $\omega$ & $k$ & Jac. Ind. & $K_\delta^{sim}$ & WNDCG & Fuzzy NMI & Fuzzy B-Cubed \\
\hline
\endhead

\hline
\endfoot

CN-ADD & - & - & 0.247 & 0.653 & 0.303 & 0.043 & 0.475 \\
CN-ADD & - & 1 & 0.227 & 0.629 & 0.289 & 0.032 & 0.478 \\
CN-ADD & - & 5 & 0.239 & 0.646 & 0.298 & 0.032 & 0.439 \\
CN-ADD & - & 15 & 0.240 & 0.637 & 0.296 & 0.035 & 0.454 \\
CN-ADD & 1 & - & 0.185 & 0.433 & 0.209 & 0.054 & 0.245 \\
CN-ADD & 2 & - & 0.252 & 0.588 & 0.293 & 0.061 & 0.373 \\
CN-ADD & 3 & - & 0.270 & 0.635 & 0.315 & 0.053 & 0.444 \\
CN-ADD & 4 & - & 0.267 & 0.637 & 0.313 & 0.055 & 0.451 \\
CN-ADD & 5 & - & 0.266 & 0.650 & 0.316 & 0.056 & 0.457 \\
CN-ADD & 7 & - & 0.260 & 0.642 & 0.313 & 0.052 & 0.461 \\
CN-ADD & 10 & - & 0.273 & 0.659 & 0.314 & 0.052 & 0.452 \\
CN-ADD & 1 & 1 & 0.220 & 0.623 & 0.283 & 0.041 & 0.469 \\
CN-ADD & 2 & 1 & 0.231 & 0.630 & 0.293 & 0.034 & 0.481 \\
CN-ADD & 3 & 1 & 0.227 & 0.628 & 0.291 & 0.034 & 0.481 \\
CN-ADD & 4 & 1 & 0.235 & 0.634 & 0.294 & 0.039 & 0.485 \\
CN-ADD & 5 & 1 & 0.237 & 0.641 & 0.293 & 0.036 & 0.483 \\
CN-ADD & 7 & 1 & 0.233 & 0.625 & 0.293 & 0.038 & 0.483 \\
CN-ADD & 10 & 1 & 0.224 & 0.633 & 0.292 & 0.035 & 0.481 \\
CN-ADD & 1 & 5 & 0.252 & 0.624 & 0.283 & 0.040 & 0.401 \\
CN-ADD & 2 & 5 & 0.257 & 0.634 & 0.294 & 0.036 & 0.409 \\
CN-ADD & 3 & 5 & 0.255 & 0.647 & 0.298 & 0.033 & 0.418 \\
CN-ADD & 4 & 5 & 0.266 & 0.651 & 0.302 & 0.037 & 0.413 \\
CN-ADD & 5 & 5 & 0.266 & 0.643 & 0.297 & 0.037 & 0.422 \\
CN-ADD & 7 & 5 & 0.260 & 0.643 & 0.299 & 0.033 & 0.423 \\
CN-ADD & 10 & 5 & 0.255 & 0.646 & 0.299 & 0.035 & 0.428 \\
CN-ADD & 1 & 15 & 0.244 & 0.606 & 0.279 & 0.041 & 0.379 \\
CN-ADD & 2 & 15 & 0.247 & 0.626 & 0.290 & 0.044 & 0.400 \\
CN-ADD & 3 & 15 & 0.257 & 0.648 & 0.307 & 0.040 & 0.431 \\
CN-ADD & 4 & 15 & 0.260 & 0.657 & 0.308 & 0.041 & 0.420 \\
CN-ADD & 5 & 15 & 0.260 & 0.646 & 0.306 & 0.039 & 0.421 \\
CN-ADD & 7 & 15 & 0.255 & 0.650 & 0.304 & 0.038 & 0.433 \\
CN-ADD & 10 & 15 & 0.257 & 0.649 & 0.302 & 0.035 & 0.425 \\
CN-AVG & - & - & 0.247 & 0.653 & 0.303 & 0.043 & 0.475 \\
CN-AVG & - & 1 & 0.227 & 0.629 & 0.289 & 0.032 & 0.478 \\
CN-AVG & - & 5 & 0.239 & 0.646 & 0.298 & 0.032 & 0.439 \\
CN-AVG & - & 15 & 0.240 & 0.637 & 0.296 & 0.035 & 0.454 \\
CN-AVG & 1 & - & 0.185 & 0.433 & 0.209 & 0.054 & 0.245 \\
CN-AVG & 2 & - & 0.257 & 0.601 & 0.302 & 0.058 & 0.404 \\
CN-AVG & 3 & - & 0.268 & 0.623 & 0.312 & 0.055 & 0.416 \\
CN-AVG & 4 & - & 0.265 & 0.633 & 0.312 & 0.055 & 0.437 \\
CN-AVG & 5 & - & 0.266 & 0.650 & 0.316 & 0.056 & 0.457 \\
CN-AVG & 7 & - & 0.260 & 0.642 & 0.313 & 0.052 & 0.461 \\
CN-AVG & 10 & - & 0.273 & 0.659 & 0.314 & 0.052 & 0.452 \\
CN-AVG & 1 & 1 & 0.220 & 0.623 & 0.283 & 0.041 & 0.469 \\
CN-AVG & 2 & 1 & 0.231 & 0.630 & 0.293 & 0.034 & 0.481 \\
CN-AVG & 3 & 1 & 0.227 & 0.628 & 0.291 & 0.034 & 0.481 \\
CN-AVG & 4 & 1 & 0.235 & 0.634 & 0.294 & 0.039 & 0.485 \\
CN-AVG & 5 & 1 & 0.237 & 0.641 & 0.293 & 0.036 & 0.483 \\
CN-AVG & 7 & 1 & 0.233 & 0.625 & 0.293 & 0.038 & 0.483 \\
CN-AVG & 10 & 1 & 0.224 & 0.633 & 0.292 & 0.035 & 0.481 \\
CN-AVG & 1 & 5 & 0.252 & 0.624 & 0.283 & 0.040 & 0.401 \\
CN-AVG & 2 & 5 & 0.257 & 0.634 & 0.294 & 0.036 & 0.409 \\
CN-AVG & 3 & 5 & 0.255 & 0.647 & 0.298 & 0.033 & 0.418 \\
CN-AVG & 4 & 5 & 0.266 & 0.651 & 0.302 & 0.037 & 0.413 \\
CN-AVG & 5 & 5 & 0.266 & 0.643 & 0.297 & 0.037 & 0.422 \\
CN-AVG & 7 & 5 & 0.260 & 0.643 & 0.299 & 0.033 & 0.423 \\
CN-AVG & 10 & 5 & 0.255 & 0.646 & 0.299 & 0.035 & 0.428 \\
CN-AVG & 1 & 15 & 0.244 & 0.606 & 0.279 & 0.041 & 0.379 \\
CN-AVG & 2 & 15 & 0.249 & 0.631 & 0.294 & 0.043 & 0.415 \\
CN-AVG & 3 & 15 & 0.257 & 0.648 & 0.307 & 0.040 & 0.431 \\
CN-AVG & 4 & 15 & 0.260 & 0.657 & 0.308 & 0.041 & 0.420 \\
CN-AVG & 5 & 15 & 0.260 & 0.646 & 0.306 & 0.039 & 0.421 \\
CN-AVG & 7 & 15 & 0.255 & 0.650 & 0.304 & 0.038 & 0.433 \\
CN-AVG & 10 & 15 & 0.257 & 0.649 & 0.302 & 0.035 & 0.425
\label{sp:tab1}
\end{longtable}

\begin{longtable}{cccccccc}
\caption{Performance of our best methods evaluated using instances that were labeled with just one sense.}\\
\hline
System & $\omega$ & $k$ & Jac. Ind. & Fuzzy NMI & Fuzzy B-Cubed \\
\hline
\endfirsthead

\caption{Performance of our best methods evaluated using instances that were labeled with just one sense (continuation).}\\
\hline
System & $\omega$ & $k$ & F1 & Fuzzy NMI & Fuzzy B-Cubed \\
\hline
\endhead

\hline
\endfoot

CN-ADD & - & - & 0.582 & 0.034 & 0.445 \\
CN-ADD & - & 1 & 0.566 & 0.023 & 0.445 \\
CN-ADD & - & 5 & 0.581 & 0.021 & 0.412 \\
CN-ADD & - & 15 & 0.574 & 0.027 & 0.425 \\
CN-ADD & 1 & - & 0.396 & 0.042 & 0.231 \\
CN-ADD & 2 & - & 0.554 & 0.049 & 0.356 \\
CN-ADD & 3 & - & 0.591 & 0.046 & 0.422 \\
CN-ADD & 4 & - & 0.592 & 0.048 & 0.426 \\
CN-ADD & 5 & - & 0.591 & 0.045 & 0.432 \\
CN-ADD & 7 & - & 0.590 & 0.044 & 0.436 \\
CN-ADD & 10 & - & 0.586 & 0.043 & 0.426 \\
CN-ADD & 1 & 1 & 0.554 & 0.034 & 0.437 \\
CN-ADD & 2 & 1 & 0.567 & 0.026 & 0.449 \\
CN-ADD & 3 & 1 & 0.568 & 0.027 & 0.449 \\
CN-ADD & 4 & 1 & 0.569 & 0.031 & 0.453 \\
CN-ADD & 5 & 1 & 0.567 & 0.027 & 0.450 \\
CN-ADD & 7 & 1 & 0.564 & 0.028 & 0.449 \\
CN-ADD & 10 & 1 & 0.561 & 0.025 & 0.447 \\
CN-ADD & 1 & 5 & 0.556 & 0.032 & 0.377 \\
CN-ADD & 2 & 5 & 0.568 & 0.028 & 0.384 \\
CN-ADD & 3 & 5 & 0.576 & 0.024 & 0.396 \\
CN-ADD & 4 & 5 & 0.577 & 0.027 & 0.390 \\
CN-ADD & 5 & 5 & 0.567 & 0.026 & 0.396 \\
CN-ADD & 7 & 5 & 0.580 & 0.023 & 0.396 \\
CN-ADD & 10 & 5 & 0.576 & 0.024 & 0.402 \\
CN-ADD & 1 & 15 & 0.525 & 0.034 & 0.357 \\
CN-ADD & 2 & 15 & 0.560 & 0.036 & 0.381 \\
CN-ADD & 3 & 15 & 0.580 & 0.033 & 0.409 \\
CN-ADD & 4 & 15 & 0.587 & 0.029 & 0.397 \\
CN-ADD & 5 & 15 & 0.590 & 0.031 & 0.399 \\
CN-ADD & 7 & 15 & 0.587 & 0.031 & 0.408 \\
CN-ADD & 10 & 15 & 0.582 & 0.028 & 0.401 \\
CN-AVG & - & - & 0.582 & 0.034 & 0.445 \\
CN-AVG & - & 1 & 0.566 & 0.023 & 0.445 \\
CN-AVG & - & 5 & 0.581 & 0.021 & 0.412 \\
CN-AVG & - & 15 & 0.574 & 0.027 & 0.425 \\
CN-AVG & 1 & - & 0.396 & 0.042 & 0.231 \\
CN-AVG & 2 & - & 0.568 & 0.049 & 0.383 \\
CN-AVG & 3 & - & 0.582 & 0.047 & 0.396 \\
CN-AVG & 4 & - & 0.591 & 0.047 & 0.415 \\
CN-AVG & 5 & - & 0.591 & 0.045 & 0.432 \\
CN-AVG & 7 & - & 0.590 & 0.044 & 0.436 \\
CN-AVG & 10 & - & 0.586 & 0.043 & 0.426 \\
CN-AVG & 1 & 1 & 0.554 & 0.034 & 0.437 \\
CN-AVG & 2 & 1 & 0.567 & 0.026 & 0.449 \\
CN-AVG & 3 & 1 & 0.568 & 0.027 & 0.449 \\
CN-AVG & 4 & 1 & 0.569 & 0.031 & 0.453 \\
CN-AVG & 5 & 1 & 0.567 & 0.027 & 0.450 \\
CN-AVG & 7 & 1 & 0.564 & 0.028 & 0.449 \\
CN-AVG & 10 & 1 & 0.561 & 0.025 & 0.447 \\
CN-AVG & 1 & 5 & 0.556 & 0.032 & 0.377 \\
CN-AVG & 2 & 5 & 0.568 & 0.028 & 0.384 \\
CN-AVG & 3 & 5 & 0.576 & 0.024 & 0.396 \\
CN-AVG & 4 & 5 & 0.577 & 0.027 & 0.390 \\
CN-AVG & 5 & 5 & 0.567 & 0.026 & 0.396 \\
CN-AVG & 7 & 5 & 0.580 & 0.023 & 0.396 \\
CN-AVG & 10 & 5 & 0.576 & 0.024 & 0.402 \\
CN-AVG & 1 & 15 & 0.525 & 0.034 & 0.357 \\
CN-AVG & 2 & 15 & 0.568 & 0.036 & 0.395 \\
CN-AVG & 3 & 15 & 0.580 & 0.033 & 0.409 \\
CN-AVG & 4 & 15 & 0.587 & 0.029 & 0.397 \\
CN-AVG & 5 & 15 & 0.590 & 0.031 & 0.399 \\
CN-AVG & 7 & 15 & 0.587 & 0.031 & 0.408 \\
CN-AVG & 10 & 15 & 0.582 & 0.028 & 0.401
\label{sp:tab2}
\end{longtable}

\begin{longtable}{cccccccc}
\caption{Performance of our best methods evaluated using instances that were labeled with multiple senses.}\\
\hline
 &  &  & \multicolumn{3}{c}{ WSD F1} & \multicolumn{2}{c}{Cluster Comparison} \\
\cline{4-6} \cline{7-8}
System & $\omega$ & $k$ & Jac. Ind. & $K_\delta^{sim}$ & WNDCG & Fuzzy NMI & Fuzzy B-Cubed \\
\hline
\endfirsthead

\caption{Performance of our best methods evaluated using instances that were labeled with multiple senses (continuation).}\\
\hline
 &  &  & \multicolumn{3}{c}{ WSD F1} & \multicolumn{2}{c}{Cluster Comparison} \\
\cline{4-6} \cline{7-8}
System & $\omega$ & $k$ & Jac. Ind. & $K_\delta^{sim}$ & WNDCG & Fuzzy NMI & Fuzzy B-Cubed \\
\hline
\endhead

\hline
\endfoot

CN-ADD & - & - & 0.441 & 0.587 & 0.260 & 0.027 & 0.131 \\
CN-ADD & - & 1 & 0.437 & 0.580 & 0.253 & 0.036 & 0.124 \\
CN-ADD & - & 5 & 0.429 & 0.572 & 0.254 & 0.022 & 0.122 \\
CN-ADD & - & 15 & 0.455 & 0.567 & 0.251 & 0.023 & 0.125 \\
CN-ADD & 1 & - & 0.288 & 0.347 & 0.161 & 0.031 & 0.064 \\
CN-ADD & 2 & - & 0.381 & 0.473 & 0.220 & 0.030 & 0.104 \\
CN-ADD & 3 & - & 0.471 & 0.540 & 0.258 & 0.024 & 0.124 \\
CN-ADD & 4 & - & 0.462 & 0.565 & 0.261 & 0.023 & 0.132 \\
CN-ADD & 5 & - & 0.455 & 0.565 & 0.256 & 0.024 & 0.130 \\
CN-ADD & 7 & - & 0.465 & 0.542 & 0.254 & 0.024 & 0.124 \\
CN-ADD & 10 & - & 0.464 & 0.562 & 0.263 & 0.021 & 0.137 \\
CN-ADD & 1 & 1 & 0.424 & 0.578 & 0.247 & 0.037 & 0.123 \\
CN-ADD & 2 & 1 & 0.458 & 0.585 & 0.255 & 0.038 & 0.127 \\
CN-ADD & 3 & 1 & 0.431 & 0.581 & 0.251 & 0.036 & 0.128 \\
CN-ADD & 4 & 1 & 0.441 & 0.595 & 0.256 & 0.040 & 0.129 \\
CN-ADD & 5 & 1 & 0.439 & 0.586 & 0.254 & 0.039 & 0.131 \\
CN-ADD & 7 & 1 & 0.438 & 0.604 & 0.257 & 0.040 & 0.131 \\
CN-ADD & 10 & 1 & 0.431 & 0.586 & 0.251 & 0.039 & 0.127 \\
CN-ADD & 1 & 5 & 0.436 & 0.555 & 0.246 & 0.019 & 0.116 \\
CN-ADD & 2 & 5 & 0.448 & 0.533 & 0.246 & 0.018 & 0.119 \\
CN-ADD & 3 & 5 & 0.448 & 0.552 & 0.249 & 0.014 & 0.119 \\
CN-ADD & 4 & 5 & 0.473 & 0.564 & 0.258 & 0.018 & 0.126 \\
CN-ADD & 5 & 5 & 0.450 & 0.568 & 0.256 & 0.020 & 0.132 \\
CN-ADD & 7 & 5 & 0.451 & 0.580 & 0.252 & 0.018 & 0.125 \\
CN-ADD & 10 & 5 & 0.461 & 0.579 & 0.259 & 0.020 & 0.129 \\
CN-ADD & 1 & 15 & 0.415 & 0.516 & 0.237 & 0.021 & 0.111 \\
CN-ADD & 2 & 15 & 0.434 & 0.520 & 0.238 & 0.016 & 0.113 \\
CN-ADD & 3 & 15 & 0.440 & 0.534 & 0.244 & 0.015 & 0.125 \\
CN-ADD & 4 & 15 & 0.454 & 0.539 & 0.249 & 0.017 & 0.117 \\
CN-ADD & 5 & 15 & 0.467 & 0.548 & 0.251 & 0.015 & 0.125 \\
CN-ADD & 7 & 15 & 0.456 & 0.558 & 0.252 & 0.022 & 0.130 \\
CN-ADD & 10 & 15 & 0.449 & 0.568 & 0.254 & 0.023 & 0.135 \\
CN-AVG & - & - & 0.441 & 0.587 & 0.260 & 0.027 & 0.131 \\
CN-AVG & - & 1 & 0.437 & 0.580 & 0.253 & 0.036 & 0.124 \\
CN-AVG & - & 5 & 0.429 & 0.572 & 0.254 & 0.022 & 0.122 \\
CN-AVG & - & 15 & 0.455 & 0.567 & 0.251 & 0.023 & 0.125 \\
CN-AVG & 1 & - & 0.288 & 0.347 & 0.161 & 0.031 & 0.064 \\
CN-AVG & 2 & - & 0.420 & 0.522 & 0.244 & 0.028 & 0.119 \\
CN-AVG & 3 & - & 0.465 & 0.533 & 0.254 & 0.025 & 0.120 \\
CN-AVG & 4 & - & 0.459 & 0.541 & 0.253 & 0.025 & 0.125 \\
CN-AVG & 5 & - & 0.455 & 0.565 & 0.256 & 0.024 & 0.130 \\
CN-AVG & 7 & - & 0.465 & 0.542 & 0.254 & 0.024 & 0.124 \\
CN-AVG & 10 & - & 0.464 & 0.562 & 0.263 & 0.021 & 0.137 \\
CN-AVG & 1 & 1 & 0.424 & 0.578 & 0.247 & 0.037 & 0.123 \\
CN-AVG & 2 & 1 & 0.458 & 0.585 & 0.255 & 0.038 & 0.127 \\
CN-AVG & 3 & 1 & 0.431 & 0.581 & 0.251 & 0.036 & 0.128 \\
CN-AVG & 4 & 1 & 0.441 & 0.595 & 0.256 & 0.040 & 0.129 \\
CN-AVG & 5 & 1 & 0.439 & 0.586 & 0.254 & 0.039 & 0.131 \\
CN-AVG & 7 & 1 & 0.438 & 0.604 & 0.257 & 0.040 & 0.131 \\
CN-AVG & 10 & 1 & 0.431 & 0.586 & 0.251 & 0.039 & 0.127 \\
CN-AVG & 1 & 5 & 0.436 & 0.555 & 0.246 & 0.019 & 0.116 \\
CN-AVG & 2 & 5 & 0.448 & 0.533 & 0.246 & 0.018 & 0.119 \\
CN-AVG & 3 & 5 & 0.448 & 0.552 & 0.249 & 0.014 & 0.119 \\
CN-AVG & 4 & 5 & 0.473 & 0.564 & 0.258 & 0.018 & 0.126 \\
CN-AVG & 5 & 5 & 0.450 & 0.568 & 0.256 & 0.020 & 0.132 \\
CN-AVG & 7 & 5 & 0.451 & 0.580 & 0.252 & 0.018 & 0.125 \\
CN-AVG & 10 & 5 & 0.461 & 0.579 & 0.259 & 0.020 & 0.129 \\
CN-AVG & 1 & 15 & 0.415 & 0.516 & 0.237 & 0.021 & 0.111 \\
CN-AVG & 2 & 15 & 0.439 & 0.525 & 0.240 & 0.016 & 0.115 \\
CN-AVG & 3 & 15 & 0.440 & 0.534 & 0.244 & 0.015 & 0.125 \\
CN-AVG & 4 & 15 & 0.454 & 0.539 & 0.249 & 0.017 & 0.117 \\
CN-AVG & 5 & 15 & 0.467 & 0.548 & 0.251 & 0.015 & 0.125 \\
CN-AVG & 7 & 15 & 0.456 & 0.558 & 0.252 & 0.022 & 0.130 \\
CN-AVG & 10 & 15 & 0.449 & 0.568 & 0.254 & 0.023 & 0.135
\label{sp:tab3}
\end{longtable}

\end{document}